\documentclass[format=acmsmall]{acmart}
\usepackage{amsmath,amsfonts}
\usepackage{algorithmic}
\usepackage{algorithm}
\usepackage{array}
\usepackage{subfigure}
\usepackage{textcomp}
\usepackage{stfloats}
\usepackage{url}
\usepackage{verbatim}
\usepackage{graphicx}
\usepackage{multirow}
\usepackage{color}
\usepackage{arydshln}
\usepackage{booktabs}
\usepackage{bbding}

\AtBeginDocument{%
  \providecommand\BibTeX{{%
    \normalfont B\kern-0.5em{\scshape i\kern-0.25em b}\kern-0.8em\TeX}}}

\setcopyright{acmlicensed}
\acmJournal{TOMM}
\acmYear{2024} \acmVolume{1} \acmNumber{1} \acmArticle{1} \acmMonth{1}\acmDOI{10.1145/3643817}

\begin{document}
\title{GMS-3DQA: Projection-based Grid Mini-patch Sampling for 3D Model Quality Assessment}
\author{Zicheng Zhang}
\affiliation{%
  \institution{Shanghai Jiao Tong University}
  \country{China}
}
\author{Wei Sun}
\affiliation{%
  \institution{Shanghai Jiao Tong University}
  \country{China}
}
\author{Haoning Wu}
\affiliation{%
  \institution{Nanyang Technological University}
  \country{Singapore}
}
\author{Yingjie Zhou}
\affiliation{%
  \institution{Shanghai Jiao Tong University}
  \country{China}
}
\author{Chunyi Li}
\affiliation{%
  \institution{Shanghai Jiao Tong University}
  \country{China}
}
\author{Zijian Chen}
\affiliation{%
  \institution{Shanghai Jiao Tong University}
  \country{China}
}
\author{Xiongkuo Min}
\affiliation{%
  \institution{Shanghai Jiao Tong University}
  \country{China}
}
\author{Guangtao Zhai}
\affiliation{%
  \institution{Shanghai Jiao Tong University}
  \country{China}
}
\author{Weisi Lin}
\affiliation{%
  \institution{Nanyang Technological University}
  \country{Singapore}
}
\renewcommand{\shortauthors}{Zhang et al.}

\begin{abstract}
{Nowadays, most 3D model quality assessment (3DQA) methods have been aimed at improving accuracy.} However, little attention has been paid to the computational cost and inference time required for practical applications. Model-based 3DQA methods extract features directly from the 3D models, which are characterized by their high degree of complexity. As a result, many researchers are inclined towards utilizing projection-based 3DQA methods. Nevertheless, previous projection-based 3DQA methods directly extract features from multi-projections to ensure quality prediction accuracy, which calls for more resource consumption and inevitably leads to inefficiency. Thus in this paper, we address this challenge by proposing a no-reference (NR) projection-based \textit{\underline{G}rid \underline{M}ini-patch \underline{S}ampling \underline{3D} Model \underline{Q}uality \underline{A}ssessment (GMS-3DQA)} method. The projection images are rendered from six perpendicular viewpoints of the 3D model to cover sufficient quality information. To reduce redundancy and inference resources, we propose a multi-projection grid mini-patch sampling strategy (MP-GMS), which samples grid mini-patches from the multi-projections and forms the sampled grid mini-patches into one quality mini-patch map (QMM). The Swin-Transformer tiny backbone is then used to extract quality-aware features from the QMMs. {The experimental results show that the proposed GMS-3DQA outperforms existing state-of-the-art NR-3DQA methods on the point cloud quality assessment databases for both accuracy and efficiency.} The efficiency analysis reveals that the proposed GMS-3DQA requires far less computational resources and inference time than other 3DQA competitors. The code is available at https://github.com/zzc-1998/GMS-3DQA.
\end{abstract}



\received{20 February 2007}
\received[revised]{12 March 2009}
\received[accepted]{5 June 2009}

\maketitle

\section{Introduction}
Three-dimensional (3D) models, such as point clouds and meshes, have garnered extensive attention and are widely utilized in communication systems, virtual/augmented reality (V/AR), auto-driving, and other applications \cite{application}. Due to the distinctive data structure and novel processing methodologies, the complexity of distortion for 3D models surpasses 2D images and 2D videos, where large amounts of quality assessment methods have been proposed \cite{gu2015quality, li2015no,yang2019modeling,zhai2021perceptual,liu2020blind,zhang2023image,zhang2021no,zhang2023md,zhang2023subjective,zhang2024quality}. Generally speaking, 3D models are frequently affected by geometry/color noise as well as compression distortion incurred during the generation and broadcasting process \cite{li2020occupancy,mekuria2016design,liu2020model,liu2021reduced}. Therefore, a reliable and efficient quality assessment metric is eagerly needed for quantifying, monitoring, and optimizing the quality of 3D-based applications.

To address these issues, numerous approaches have been proposed to evaluate the visual quality of degraded 3D models, known as 3D model quality assessment (3DQA) methods.
3DQA methods can be broadly divided into two categories: model-based and projection-based methods. Model-based 3DQA methods involve the direct extraction of features from 3D models, which are known for their intricacy. Therefore, researchers lean towards the utilization of projection-based 3DQA methods. Unlike model-based 3DQA methods that extract features directly from the 3D models, projection-based methods evaluate the quality of 3D models through rendered 2D projections. This kind of approach capitalizes on mature 2D vision technology to achieve cost-effective performance. However, projections are highly dependent on the viewpoint and a single projection may not cover sufficient quality information. To tackle this challenge, many projection-based 3DQA methods \cite{yang2020predicting,liu2021pqa,fan2022no,zhang2022treating,10091221,zhang2023perceptual,zhang2023advancing} employ multi-projections to help improve performance as well as robustness and it has been firmly proven that using multi-projections is superior to using single projection. Unfortunately, the multi-projection strategy often entails additional rendering time and high computational demands. 

\begin{figure}
    \centering
    \includegraphics[width = 0.7\linewidth]{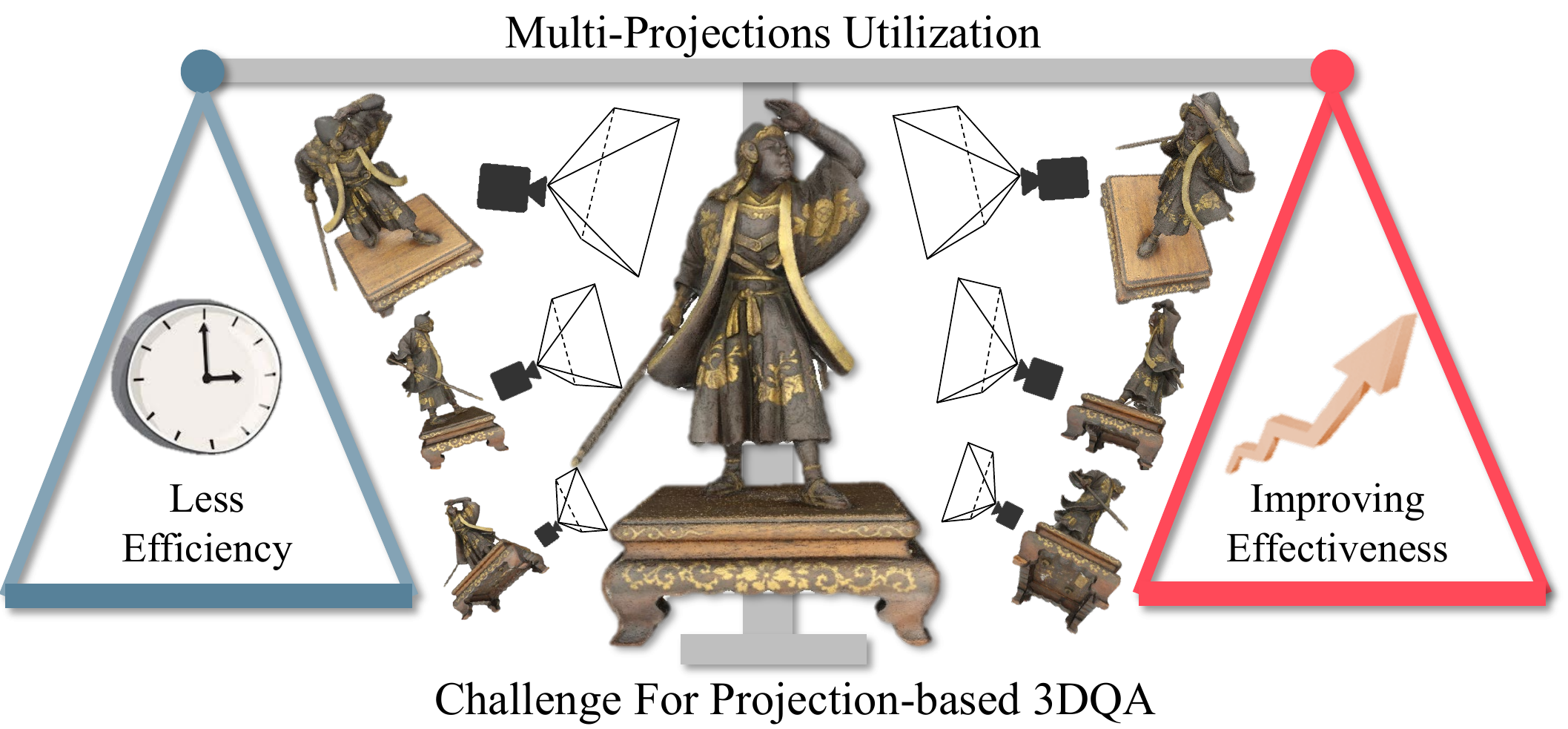}
    \caption{Illustration of the challenge for the projection-based 3DQA methods, where employing multi-projections can help improve the performance but inevitably consume more computational resources.}
    \label{fig:spotlight}
    \vspace{-0.4cm}
\end{figure}

To guarantee efficiency and effectiveness, the core contradiction (shown in Fig. \ref{fig:spotlight}) that needs to be resolved is \textbf{how to use the multi-projection information while reducing the inference time as much as possible.} Therefore, in this paper, we propose a projection-based \textit{\underline{G}rid \underline{M}ini-patch \underline{S}ampling \underline{3D} Model \underline{Q}uality \underline{A}ssessment (GMS-3DQA)} method to deal with this problem. Specifically, six perpendicular viewpoints are fixed for rendering projections, which have been adopted as the common rendering setup in many 3DQA-related works \cite{yang2020predicting,graziosi2020overview}. The computational cost of rendering these projections is within acceptable limits, and the predefined viewpoints eliminate the need for additional time spent on viewpoint selection, thereby enhancing efficiency. Inspired by the success of grid mini-patch sampling proposed in \cite{wu2022fast,wu2022neighbourhood,wu2022disentangling}, we split the projections into grid mini-patch maps, which serve to depict the quality patterns of the 3D models. Afterward, we randomly sample mini-patches from each viewpoint and splice the mini-patches into one quality mini-patch map (QMM) for evaluation. The use of a single QMM to assess the quality of the 3D models instead of extracting features from each projection results in improved efficiency (as detailed in Section \ref{sec:efficiency}). The QMM has the added benefit of aggregating quality information across different viewpoints and mitigating the negative effects of information redundancy among the projections, thereby improving both performance and robustness (as discussed in Sections \ref{sec:performance} and \ref{sec:cross}). The Swin Transformer \cite{liu2021swin} has a hierarchical structure and processes inputs with patch-wise operations, therefore it is naturally suitable for processing grid mini-patch maps \cite{wu2022fast}. Then the Swin Transformer tiny is used as the backbone and the fully-connected layers are used to regress the features into quality scores. 

The proposed GMS-3DQA is validated on both point cloud quality assessment (PCQA) and mesh quality assessment (MQA) tasks, of which the validated databases include the SJTU-PCQA \cite{yang2020predicting}, WPC \cite{liu2022perceptual}, CMDM \cite{y2021visual} and TMQ \cite{nehme2022textured}. The experimental results show that GMS-3DQA outperforms all the comparing NR-3DQA methods on the SJTU-PCQA, WPC, and TMQ databases, \textbf{surpassing the second-ranked method by margins of 7\%, 4.2\%, and 45.6\% respectively}. The efficiency analysis demonstrates that \textbf{GMS-3DQA is 3.28 times faster than the fastest comparing 3DQA methods on CPU with fewer parameters and FLOPs}. The cross-database validation demonstrates the generalized capabilities and stability of GMS-3DQA, showing that it is able to handle unseen 3D content with different captured methods. Furthermore, the cross-domain evaluation highlights its potential to address QA tasks in a target 3D domain that lacks sufficient annotations, by leveraging the knowledge acquired from other related 3D domains (represented in different digital formats) with adequate labeling. Our contributions are listed as follows:
\begin{itemize}
    \item We propose a novel multi-projection grid mini-patch sampling (MP-GMS) strategy for the 3DQA methods, which samples mini-patches from multi-projections and forms the mini-patches into one quality map for evaluation. 
    \item The MP-GMS strategy enables the proposed method to effectively and efficiently evaluate the perceptual quality of 3D models, leading to at least 4.2\% performance margin and notable efficiency (at least 3.28 times faster) than the existing 3DQA methods. 
    \item Extensive experimental results (efficiency analysis, ablation study, cross-database \& cross-domain validation, and statistical test) have validated the rationality and robustness of the proposed method framework. In-depth analyses are further given to support the findings.
\end{itemize}

The rest paper is organized as follows: Section \ref{sec:related} briefly reviews the development of PCQA and MQA. Section \ref{sec:proposed} describes the technical details of the proposed method. Section \ref{sec:experiment} presents the experimental results of the proposed methods and the comparing methods. Section \ref{sec:conclusion} concludes this paper.

\section{Related Works}
\label{sec:related}
In this section, we briefly summarize the development of 3DQA methods from the aspects of PCQA and MQA.

\subsection{PCQA development}
The early FR-PCQA metrics primarily focused on evaluating the geometry aspect at the point level, such as p2point \cite{mekuria2016evaluation}, p2plane \cite{tian2017geometric}. The p2point metric calculates the level of distortion by determining the distance vector between corresponding points, while the p2plane metric extends this by projecting the distance vector onto the normal orientation for quality assessment. 
Due to the difficulty in reflecting complex structural distortions through point-level differences, subsequent studies considered other structural characteristics of PCQA. For instance, the angular difference of point normals was adopted by Alexiou $et$ $al$. \cite{alexiou2018point} to estimate degradation, while Javaheri $et$ $al$. \cite{javaheri2020genralized} utilized the generalized Hausdorff distance to reflect the impact of compression operations.
In some cases, the color information cannot be disregarded, which poses a challenge for PCQA metrics that only consider geometry information. To address this, PSNR-yuv \cite{torlig2018novel} measures the quality levels by comparing the point-wise color attributes. GraphSIM \cite{yang2020graphsim} predicts the quality of point clouds through the use of graph similarity and color gradient analysis. Meynet $et$ $al$. \cite{meynet2020pcqm} proposed a metric that incorporates color information by using a weighted linear combination of curvature and color information to evaluate the visual quality of distorted point clouds. Similarly, PointSSIM \cite{alexiou2020pointssim} computes the similarity of four types of features, including geometry, normal vectors, curvature, and color information. In addition, some studies have attempted to assess the visual quality of point clouds through 2D projections. For example, the works presented in \cite{yang2020predicting} and \cite{torlig2018novel} utilize mature image quality assessment (IQA) methods to evaluate the point cloud quality by projecting the 3D data onto 2D images.

Recently, more efforts have been put into pushing forward the development of NR-PCQA methods. 
ResCNN \cite{liu2022point} proposes an end-to-end sparse convolutional neural network architecture for the assessment of point cloud quality through the extraction of quality-sensitive features. PQA-net \cite{liu2021pqa} utilizes multi-view projections to classify and evaluate point cloud distortions. 3D-NSS \cite{zhang2022no} employs a combination of geometry and color attributes and implements classical statistical distribution models to analyze the point cloud. 
Tu $et$ $al.$ \cite{tu2022v} designed a dual-stream convolutional network to extract the texture and geometry features of the distorted point clouds. Zhou $et$ $al.$ \cite{zhou2022blind} extracts quality-related features with structure-guided resampling. Moreover, some works \cite{fan2022no,zhang2022treating} transform the point clouds into videos and assess the perceptual quality with the video quality assessment (VQA) technique. 

{Based on the above review, PCQA methods can be divided into two main categories: model-based methods~\cite{mekuria2016evaluation,tian2017geometric,alexiou2018point,javaheri2020genralized,yang2020graphsim,meynet2020pcqm,alexiou2020pointssim,torlig2018novel,liu2022point,zhou2022blind}, which extract features directly from the point cloud, and projection-based methods~\cite{yang2020predicting,liu2021pqa,fan2022no,zhang2022treating,xie2023pmbqa,zhang2023mm}, which extract features from rendered projections. The model-based methods avoid information loss during rendering but require significant computational resources due to the complexity of high-quality point clouds. Conversely, projection-based methods leverage mature 2D IQA tools but depend heavily on the choice of viewpoint. To reduce the randomness in viewpoint selection, it has been demonstrated in various studies \cite{liu2021pqa, fan2022no, zhang2022treating} that using multiple projections can greatly enhance accuracy compared to a single projection. However, this approach increases computational resource usage and inference time. Thus, improving computational efficiency while relying on multi-projection for enhanced performance is an urgent issue that needs addressing.}

\begin{figure*}
    \centering
    \includegraphics[width = .98\linewidth]{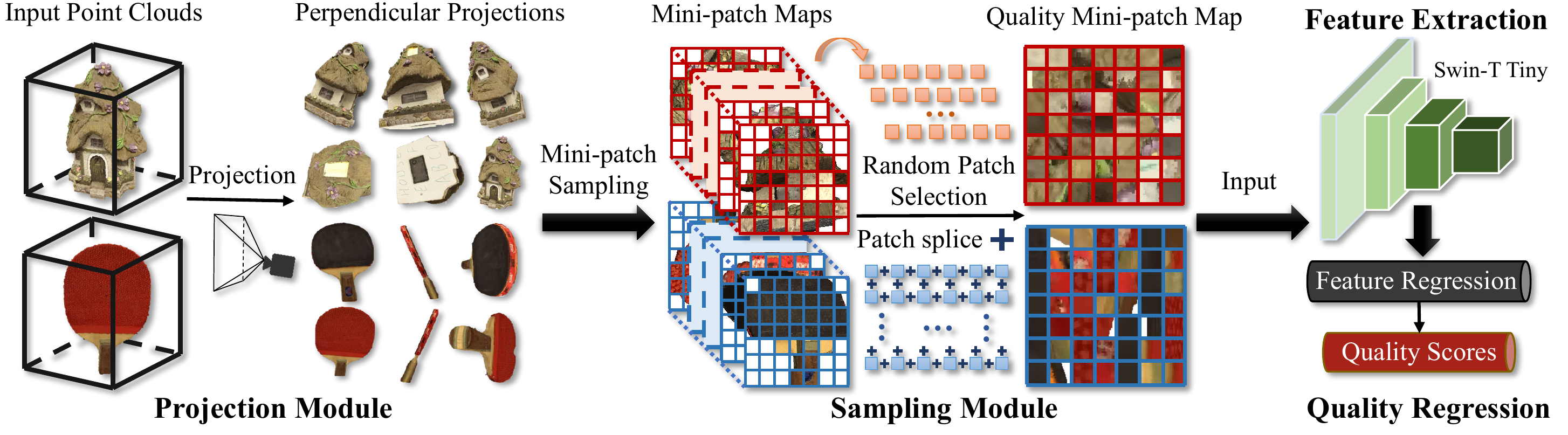}
    \caption{The framework of the proposed method. Perpendicular projections are captured from the input point clouds corresponding to the six surfaces of the cube, which are further processed into mini-patch maps. Patches are randomly selected from the mini-patch maps and then spliced into quality mini-patch maps (QMMs) for evaluation. Finally, the quality-aware features are extracted by Swin-Transformer tiny and regressed into perceptual scores.}
    \label{fig:framework}
    \vspace{-0.4cm}
\end{figure*}

\begin{figure}
    \centering
    \includegraphics[width = .98\linewidth]{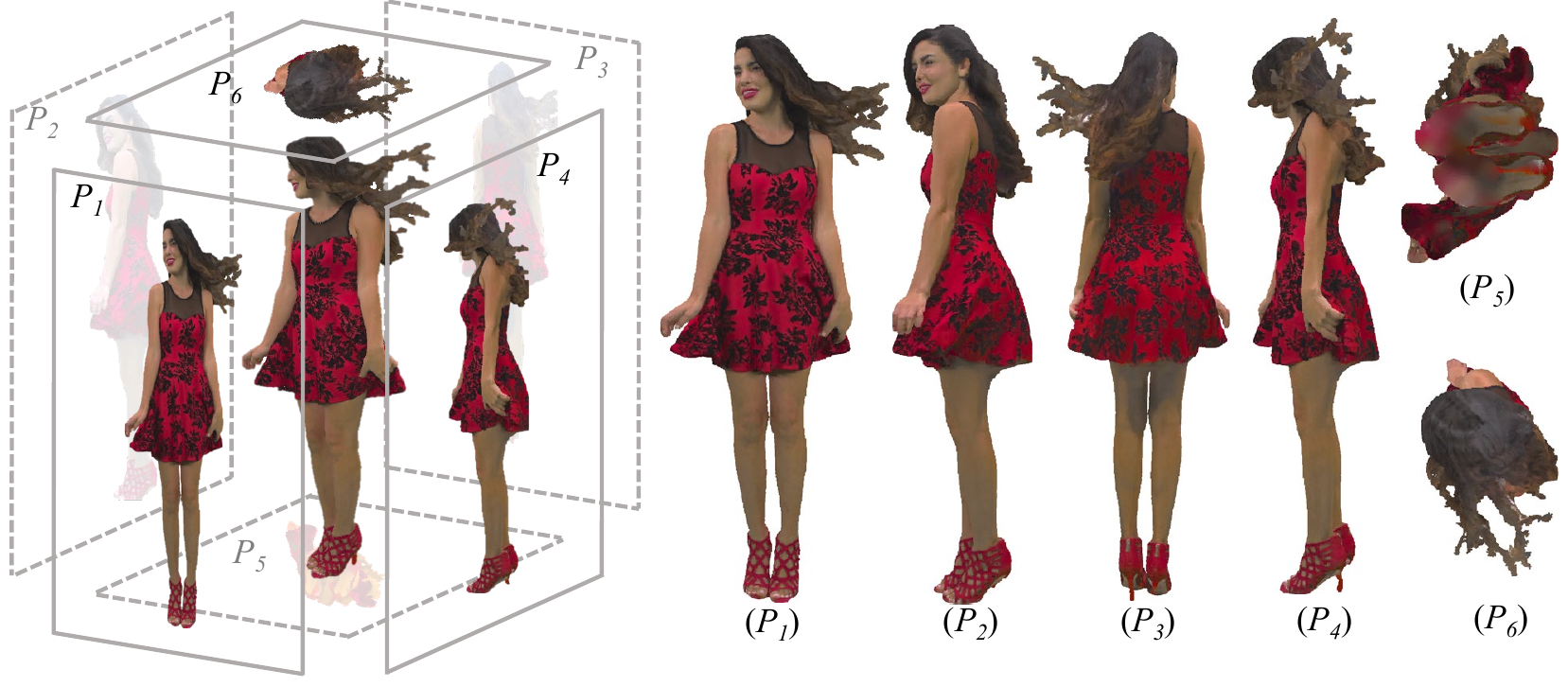}
    \caption{{Illustration of the projection process, where the projections are rendered from the cube-like six perpendicular viewpoints.}}
    \label{fig:projection}
    \vspace{-0.4cm}
\end{figure}

\subsection{MQA development}
Similar to the early FR-PCQA methods, some FR-MQA methods generally compute local features at the vertex level and then aggregate these features into a quality value. For instance, MSDM2 \cite{m1} predicts the quality level by computing the differences in structure, as captured through curvature statistics, in local neighborhoods. DAME \cite{dame} evaluates the quality loss by measuring differences in dihedral angles between the reference and distorted meshes. FMPD \cite{ff2_roughness} assesses the quality of distorted meshes by estimating the local roughness difference derived from Gaussian curvature. However, it is worth noting that these metrics only take geometry information into account.
To account for the impact of color information, some color-involved FR-MQA metrics have been developed. Tian $et$ $al.$ \cite{tian-color} quantified the effect of color information through a global distance over the texture image, calculated using Mean Squared Error (MSE). Guo $et$ $al.$ \cite{guo-color} calculated the texture image quality distance as a representation of color information features. Nehmé $et$ $al.$ \cite{y2021visual} introduced a metric that incorporates perceptually relevant curvature-based and color-based features to evaluate the visual quality of colored meshes. Then Nehmé $et$ $al.$ \cite{nehme2022textured} further proposed a learning-based quality metric for textured meshes based on LPIPS \cite{zhang2018unreasonable}. 

More recently, the efficacy of machine learning technologies has enabled the development of learning-based NR-MQA metrics. Abouelaziz $et$ $al.$ \cite{nr-svr} extracted features using dihedral angle models and train a support vector machine for feature regression.  Later, Abouelaziz $et$ $al.$ \cite{nr-cnn} convert the curvature and dihedral angle into 2D patches and employ a Convolutional Neural Network (CNN) for training. They further introduce a CNN framework with saliency views rendered from 3D meshes \cite{nr-cnncmp}. To further deal with colored meshes, Zhang $et$ $al.$ \cite{zhang2021mesh} employed curvature and color features for quality prediction.

\section{Proposed Method}
\label{sec:proposed}
The framework of the proposed method is illustrated in Fig. \ref{fig:framework}, which includes the projection module, sampling module, feature extraction module, and quality regression module. { The first stage involves capturing perpendicular projections from the six surfaces of a cube, using these projections from the input point clouds. These projections are then transformed into mini-patch maps. From these maps, patches are randomly chosen and assembled into Quality Mini-Patch Maps (QMMs) for assessment. The final stage of the process involves extracting quality-aware features using a Swin-Transformer tiny model, which are then used to regress into perceptual scores for evaluation.}

\subsection{Projection Process}
The projection-based 3DQA methods can be adapted to all kinds of 3D models, i.e., point cloud, mesh, voxel, etc., since they infer the visual quality via the rendered projections. However, the quality information contained in the projections is highly dependent on the viewpoints. To cover sufficient quality information across different viewpoints, we employ the mainstream cube-like viewpoints setting, which has been employed in the popular point cloud compression standard MPEG VPCC \cite{graziosi2020overview}. As shown in Fig. \ref{fig:projection}, six perpendicular viewpoints are employed to capture the rendered projections, corresponding to the six surfaces of a cube. Given a 3D model $\mathbf{O}$, the projection process can be described as:

\begin{equation}
\begin{aligned}
     &\mathbf{P}  = \psi(\mathbf{O}), \\
    \mathbf{P}  = \{&\mathcal{P}_{k}|k =1,\cdots, 6\},
\end{aligned}
\end{equation}
where $\mathbf{P}$ represents the set of the 6 rendered projections and $\psi(\cdot)$ stands for the rendering process.

\begin{figure*}
    \centering
    \includegraphics[width = \linewidth]{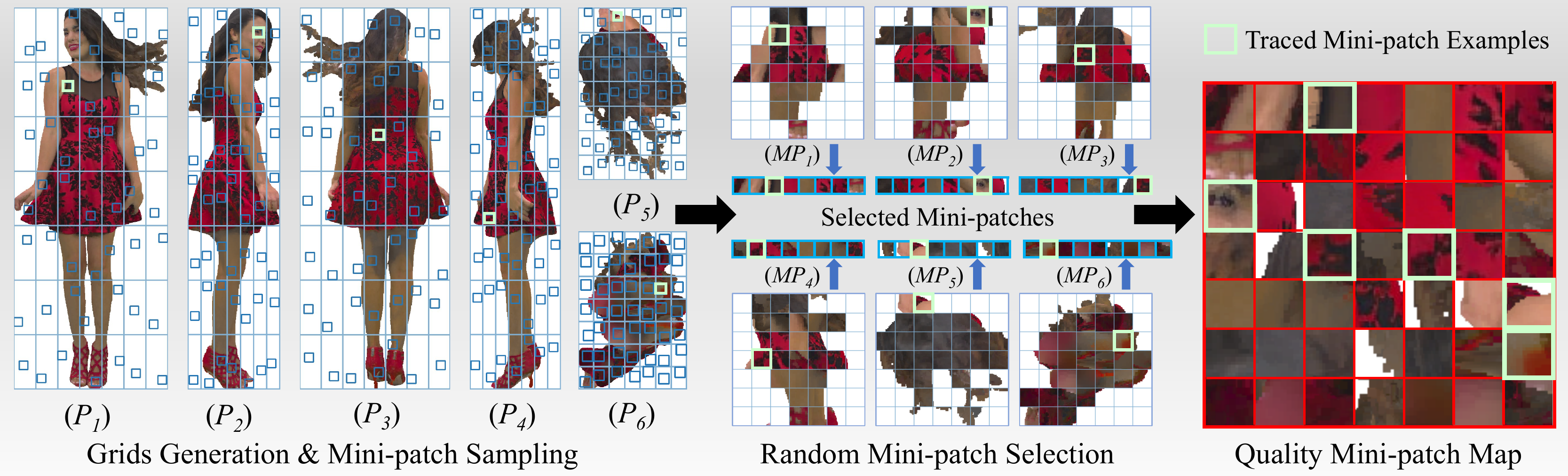}
    \caption{An example of the multi-projection grid mini-patch sampling process. $\mathcal{L} \times \mathcal{L}$ ($7 \times 7$ for this example) uniform grids are generated from the multi-projections and mini-patches are sampled from the uniform grids. Then $\lfloor \mathcal{L}^2/6 \rfloor$ (8 for this example) mini-patches are randomly sampled from each projection's mini-patch map and the last mini-patch map provides 1 extra mini-patch to fill up the QMM. It's worth mentioning that the blank mini-patches are ignored.}
    \label{fig:mini_patch}
\end{figure*}

\begin{figure}[!t]
    \centering
    \subfigure[Noise]{\begin{minipage}[t]{0.44\linewidth}
                \centering
                \includegraphics[width = 0.94\linewidth,height = 0.48\linewidth]{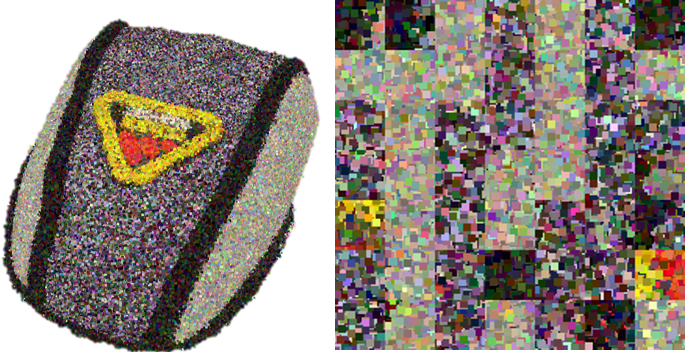}
                \end{minipage}}
    \subfigure[Downsampling]{\begin{minipage}[t]{0.44\linewidth}
                \centering
                \includegraphics[width = 0.96\linewidth,height = 0.48\linewidth]{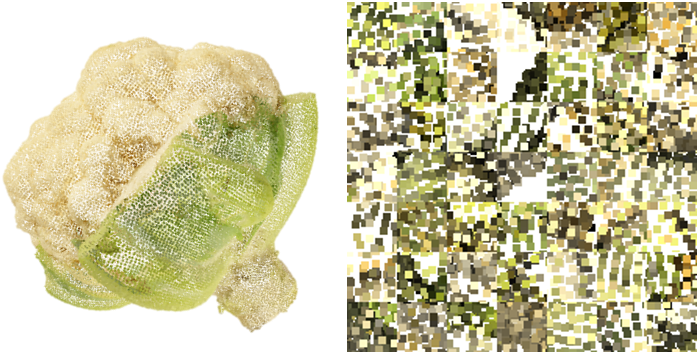}
                \end{minipage}}
    \subfigure[VPCC Compression]{\begin{minipage}[t]{0.44\linewidth}
                \centering
                \includegraphics[width = 0.94\linewidth,height = 0.48\linewidth]{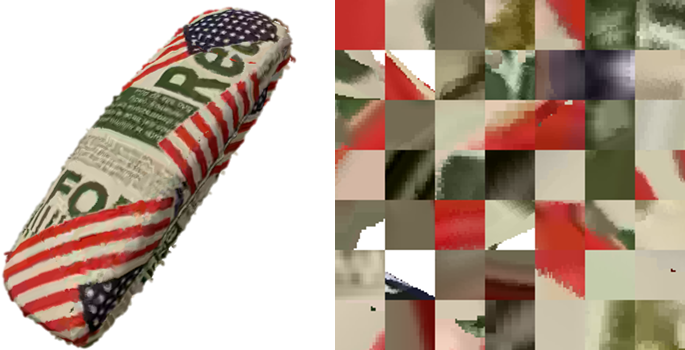}
                \end{minipage}}
    \subfigure[GPCC Compression]{\begin{minipage}[t]{0.44\linewidth}
                \centering
                \includegraphics[width = 0.96\linewidth,height = 0.48\linewidth]{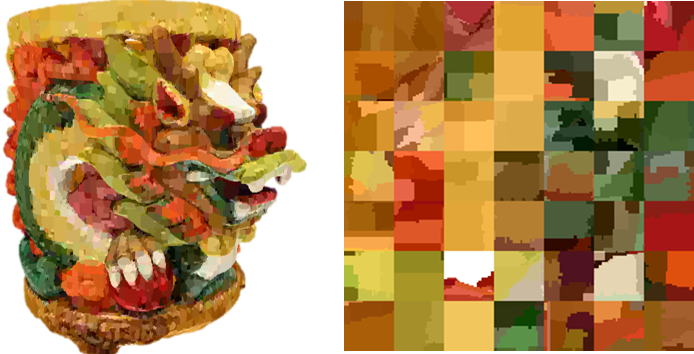}
                \end{minipage}}
    \caption{Examples of the distorted point clouds and their corresponding QMMs, where we can see the noise and the patterns are well-preserved and even more obvious in the QMMs. The VPCC and GPCC compression standards are issued by the MPEG group for point cloud compression \cite{schwarz2018emerging}. More specifically, the VPCC compression introduces blur to the mini-patches while the GPCC compression causes more artifacts as well as block effect. }
    \label{fig:distortion}
    \vspace{-0.4cm}
\end{figure}

\subsection{Multi-Projection Grid Mini-patch Sampling}
\label{sec:sampling}
It has been proven that utilizing multi-projections can help projection-based 3DQA methods learn better quality representations and  boost performance \cite{yang2020predicting,liu2021pqa,fan2022no,zhang2022treating}. 
However, such methods simply extract features from the multi-projections with deep neural networks (DNN) for quality evaluation, and using more projections means more computational resources as well as inference time. Moreover, there usually exists quality information redundancy among the multi-projections. To leverage the advantages of multi-projections and reduce the computation consumption at the same time, we propose a multi-projection grid mini-patch sampling (MP-GMS) strategy, which splits the multi-projections into grid mini-patches and forms the sampled grid mini-patches into one quality mini-patch map (QMM). 
Specifically, the grids from the multi-projections can be derived as:

\begin{equation}
\begin{aligned}
    & \mathcal{G}_{k} = \{g_{0,0}^k,\cdots, g_{i,j}^k,\cdots,g_{\mathcal{L}-1,\mathcal{L}-1}^k\},\\
    g^k_{i, j}=\mathcal{P}_k  [&\frac{i \!\times\! H_k}{\mathcal{L}}\!:\! \frac{(i\!+\!1) \!\times\! H_k}{\mathcal{L}}, \frac{j \!\times\! W_k}{\mathcal{L}}\!: \!\frac{(j\!+\!1) \!\times\! W_k}{\mathcal{L}}], 
\end{aligned}
\end{equation}
where $\mathcal{G}_{k}$ represent the uniform $\mathcal{L} \times \mathcal{L}$ grids generated from the $k$-th projection ${P}_k$,  $g^k_{i, j}$ denotes the grid in the $i$-th row and $j$-th column of ${P}_k$, and $H_k$ and $W_k$ denote the height and width of ${P}_k$. Then the raw resolution mini-patches can be obtained as: 

\begin{equation}
     \mathcal{MP}_{i,j}^k = \Theta(g^k_{i, j}),
\end{equation}
where $\mathcal{MP}_{i,j}^k$ is the mini-patch sampled from grid $g^k_{i, j}$ and $\Theta(\cdot)$ represents the patch sampling operation. Afterward, we randomly select $N_{\alpha}$ mini-patches from each projection and splice the selected mini-patches into one QMM:

\begin{equation}
    \mathcal{QM} = \mathop{\oplus}\limits_{k=1}^6 \mathop{\oplus}\limits_{\alpha=1}^{N_{\alpha}} \mathcal{MP}_{i_{\alpha},j_{\alpha}}^k
\end{equation}
where $\mathcal{QM}$ represents the QMM, $\oplus$ indicates the mini-patch splicing operation, and $\mathcal{MP}_{i_{\alpha},j_{\alpha}}^k$ stands for the randomly selected mini-patch sampled from the grid in the $i_\alpha$-th row and $j_\alpha$-th column of the $k$-th projection where the blank mini-patches are ignored. Additionally, $N_\alpha$ is set as $\lfloor \mathcal{L}^2/6 \rfloor$, which indicates that $\lfloor \mathcal{L}^2/6 \rfloor$ mini-patches are randomly selected from each projection. The left $\mathcal{L}^2 - 6\lfloor \mathcal{L}^2/6 \rfloor$ mini-patches of the quality map are filled up by the last projection. The detailed process is exhibited in Fig. \ref{fig:mini_patch}, from which we can clearly observe the generation process of QMMs. Several distorted point clouds and their corresponding QMMs are exhibited in Fig \ref{fig:distortion}, from which we can find that the multi-projections' quality-aware local patterns are well preserved in the QMM.

\subsection{Efficient Feature Extraction \& Quality Regression}
\label{sec:regression}
To ensure efficiency and suit the mini-patch structure, we choose the lightweight and well-performing Swin-Transformer tiny (ST-t) \cite{liu2021swin} as the feature extraction backbone. Given the input quality mini-patch map $\mathcal{QM}$, the quality representation can be derived as:
\begin{equation}
    F_{\mathcal{QM}} = \mathcal{ST}(\mathcal{QM}), 
\end{equation}
where $\mathcal{ST}(\cdot)$ stands for the feature extraction operation with the ST-t backbone and $F_{\mathcal{QM}}$ denotes the extracted quality representation. Then we simply use two-stage fully-connected (FC) layers with 768 and 64 neurons to regress the quality representation into predicted perceptual quality scores:
\begin{equation}
    Q = \mathcal{FC}(F_{\mathcal{QM}}), 
\end{equation}
where $Q$ represents the predicted quality scores and $\mathcal{FC}(\cdot)$ indicates the quality regression operation with the FC layers.

\subsection{Loss Function}
In common situations, we not only pay attention to the accuracy of predicted quality scores but also focus on the quality rankings for the quality assessment tasks \cite{fang2022perceptual,gu2019no}. To ensure the accuracy of quality prediction and quality rankings, the loss function employed in this paper is made up of two parts: Mean Squared Error (MSE) and Rank Error (RE). The MSE is used to force the predicted quality scores close to the quality labels, which can be described as:
\begin{equation}
    L_{MSE} = \frac{1}{n} \sum_{\eta=1}^{n} (Q_{\eta}-Q_{\eta}')^2,
\end{equation}
where $Q_{\eta}$ represents the predicted quality scores, $Q_{\eta}'$ indicates the quality labels, and $n$ is the size of a batch. Furthermore, the RE can assist the model to gain a better understanding of the quality rankings and distinguish the quality difference of point clouds with close quality labels. Specifically, we use the differentiable rank function described in \cite{wen2021strong,sun2022deep} to approximate the RE loss:

\begin{equation}
\begin{aligned}
    L_{RE}^{a b}\!=\!\max\! &\left(0,\left|Q_{a}-Q_{b}\right|\!-\!e\left(Q_{a}, Q_{b}\right)\! \cdot\! \left({Q}_{a}'-{Q}_{b}'\right)\right), \\
    &e\left(Q_{a}, Q_{b}\right)=\left\{\begin{array}{r}
1, Q_{a} \geq Q_{b}, \\
-1, Q_{a}<Q_{b},
\end{array}\right.
\end{aligned}
\end{equation}
where $a$ and $b$ are the corresponding indexes for two point clouds in a mini-batch and the RE loss is obtained as:
\begin{equation}
    L_{RE}=\frac{1}{n^{2}} \sum_{a=1}^{n} \sum_{b=1}^{n} L_{RE}^{a b},
\end{equation}
Then the overall loss is formulated as a weighted linear combination of MSE loss and RE loss:
\begin{equation}
    Loss=\lambda_{1}L_{MSE}+\lambda_{2} L_{RE}
\end{equation}
where $\lambda_{1}$ and $\lambda_{2}$ are used to control the proportion of the MSE loss and the RE loss.

\section{Experiment}
\label{sec:experiment}
In this section, we first briefly introduce the benchmark databases and the implementation details. The experimental results are presented with in-depth discussions.

\subsection{Databases}
\begin{itemize}
    \item PCQA Databases: Two popular point cloud quality assessment databases are utilized to validate the performance of the proposed method, which include the subjective point cloud assessment database (SJTU-PCQA) \cite{yang2020predicting}, the Waterloo point cloud assessment database (WPC) \cite{liu2022perceptual}. The SJTU-PCQA database includes 9 reference point clouds, each of which is corrupted with 7 types of distortions (compression, color noise, geometry noise, downsampling, and their three combinations) under 6 different strengths, resulting in a total of 378 distorted point clouds. The WPC database contains 20 reference point clouds. Each of the reference point clouds is augmented with 4 types of distortions (downsampling, Gaussian noise, MPEG-GPCC compression, and MPEG-VPCC compression), generating 740 distorted point clouds.
    
    \item MQA Databases: The diffuse color mesh quality assessment (CMDM) and the textured mesh quality (TMQ) database \cite{nehme2022textured} are used to validate the performance on the mesh quality assessment task.
 The CMDM database includes 5 source color meshes and degrades the color meshes with 4 types of distortions (geometric quantization, color quantization, ”Color-ignorant” simplification, and ”Color-aware” simplification) under 4 strengths, which generates 80 distorted stimuli in total. The TMQ database provides 55 source textured meshes and corrupts the source textured meshes with 5 common distortions (simplification, position quantization, UV map quantization, texture downsampling, and texture compression), in which 3,000 distorted textured meshes are selected and perceptually rated. 
\end{itemize}

\subsection{Implementation Details}
The Adam optimizer \cite{kingma2014adam} is employed with the $1\times{10}^{-4}$ initial learning rate and the learning rate decays with a ratio of 0.9 for every 5 epochs. The default batch size is set as 32 and the default training epochs are set as 50. 
The Swin-Transformer tiny \cite{liu2021swin} is initialized with the weights pretrained on the ImageNet-22K database \cite{deng2009imagenet}. The projections are acquired with the assistance of Open3d \cite{open3d}. The parameter $\mathcal{L}$ introduced in Section \ref{sec:sampling} is set as 7, which indicates $7 \times 7$ grids are generated from each projection and the final QMM is formed with $7 \times 7$ selected mini-patches. 
The $\lambda_{1}$ and $\lambda_{2}$ parameters described in Section \ref{sec:regression} are both set as 1 respectively. 

\subsection{Quality Competitors \& Criteria}
\begin{itemize}
    \item { PCQA Competitors: The compared FR quality assessment methods include MSE-p2point (MSE-p2po) \cite{mekuria2016evaluation}, Hausdorff-p2point (HD-p2po) \cite{mekuria2016evaluation}, MSE-p2plane (MSE-p2pl) \cite{tian2017geometric}, Hausdorff-p2plane (HD-p2pl) \cite{tian2017geometric}, PSNR-yuv \cite{torlig2018novel}, PCQM \cite{meynet2020pcqm}, GraphSIM \cite{yang2020graphsim}, PointSSIM \cite{alexiou2020pointssim}, PSNR, and SSIM \cite{ssim}. The compared NR methods include 3D-NSS \cite{zhang2022no}, ResSCNN \cite{liu2022point}, GPA-net \cite{shan2023gpa}, PQA-net \cite{liu2021pqa}, IT-PCQA \cite{yang2022no}, and VQA-PC \cite{zhang2022treating}.}
    
    \item MQA Competitors: The compared FR quality assessment methods include PSNR, SSIM \cite{ssim}, and G-LPIPS (specially designed for textured meshes) \cite{nehme2022textured}. The compared NR methods include 3D-NSS \cite{zhang2022no}, BRISQUE \cite{mittal2012brisque}, and NIQE \cite{mittal2012making}.
\end{itemize}

It's worth noting that PSNR, SSIM, BRISQUE, and NIQE are operated on all 6 projections and the average scores are recorded. 
Afterward, a five-parameter logistic function is applied to map the predicted scores to subjective ratings, as suggested by \cite{antkowiak2000final}:
\begin{equation}
\hat{y}=\beta_{1}\left(0.5-\frac{1}{1+e^{\beta_{2}\left(y-\beta_{3}\right)}}\right)+\beta_{4} y+\beta_{5},
\end{equation}
where $\left\{\beta_{i} \mid i=1,2, \ldots, 5\right\}$ are parameters to be fitted, $y$ and $\hat{y}$ are the predicted scores and mapped scores respectively.

The evaluation of the performance of 3DQA quality models is done using four mainstream criteria, which include Spearman Rank Correlation Coefficient (SRCC), Pearson Linear Correlation Coefficient (PLCC), Kendall’s Rank Order Correlation Coefficient (KRCC), and Root Mean Squared Error (RMSE). SRCC gauges the correlation of ranks, PLCC represents linear correlation, KRCC reflects the likeness of the orderings, while RMSE measures the quality prediction accuracy. A top-performing model should have SRCC, PLCC, and KRCC values that approach 1 and RMSE values close to 0.

\subsection{Experimental Setup}
To fully evaluate the performance of the quality models, we use the $k$-fold cross validation strategy. Specifically, we divide the database into $k$ equally sized folds. The model is then trained on $k$-1 of these folds and tested on the remaining one. This process is repeated k times, each time using a different fold as the test set until each fold has been used as the test set once. 
The average performance of $k$-fold cross validation is reported as the final performance to avoid randomness. 

Considering that the SJTU-PCQA, WPC, CMDM and TMQ databases contain 9, 20, 5, and 55 independent groups of 3D models, we conduct the $9$-fold, $5$-fold, $5$-fold, and $5$-fold cross validation on the four databases respectively, which indicates each fold contains 1, 4, 1, and 11 groups of 3D models for the four databases.
It's worth mentioning that there is no content overlap between the training and testing sets. Furthermore, the quality models that require no training are evaluated on the same testing folds and the average performance is reported to make the comparison fair.

\begin{table*}[!tp]
\centering 
\renewcommand\tabcolsep{4pt}
\caption{Performance results on the SJTU-PCQA and WPC databases. The best performance results are marked in {\bf\textcolor{red}{RED}} and the second performance results are marked in {\bf\textcolor{blue}{BLUE}}.}
\resizebox{\linewidth}{!}{\begin{tabular}{l:l:c:l|cccc|cccc}
\toprule
\multirow{2}{*}{Ref}&\multirow{2}{*}{Type}&\multirow{2}{*}{Index}&\multirow{2}{*}{Methods} & \multicolumn{4}{c|}{SJTU-PCQA} & \multicolumn{4}{c}{WPC} \\ \cline{5-12}
        &&& & SRCC$\uparrow$      & PLCC$\uparrow$      & KRCC$\uparrow$     & RMSE $\downarrow$    & SRCC$\uparrow$      & PLCC$\uparrow$      & KRCC$\uparrow$       & RMSE $\downarrow$ \\ \hline
\multirow{10}{*}{FR} &\multirow{8}{38pt}{{{Model-based}}} 
 &A&MSE-p2po & 0.7294 & 0.8123 & 0.5617 & 1.3613 & 0.4558 & 0.4852 & 0.3182 & 19.8943 \\
 &&B&HD-p2po & 0.7157 & 0.7753 & 0.5447 & 1.4475 &0.2786 &0.3972&0.1943 &20.8990\\
 &&C&MSE-p2pl & 0.6277 & 0.5940 & 0.4825 & 2.2815 & 0.3281 & 0.2695 &0.2249 & 22.8226 \\
 &&D&HD-p2pl & 0.6441   & 0.6874    & 0.4565    & 2.1255 & 0.2827 & 0.2753 &0.1696 &21.9893 \\
 &&E&PSNR-yuv & 0.7950 & 0.8170 & 0.6196 & 1.3151 & 0.4493 & 0.5304 & 0.3198 & 19.3119\\
 &&F& PCQM        & {0.8644}   & {0.8853}    & \bf\textcolor{blue}{0.7086}     & {1.0862}     &{0.7434}    & {0.7499}   & {0.5601}   & 15.1639             \\
&&G& GraphSIM    & \bf\textcolor{blue}{0.8783}    & 0.8449    & {0.6947}   & \bf\textcolor{blue}{1.0321}  & 0.5831    & 0.6163    & 0.4194   & 17.1939   \\
&&H& PointSSIM      & 0.6867  & 0.7136  & 0.4964 & 1.7001  & 0.4542    & 0.4667    & 0.3278   & 20.2733  \\ \cdashline{2-12}
&\multirow{2}{38pt}{{{Projection-based}}} &I&
PSNR &0.2952 &0.3222 &0.2048 &2.2972  &0.1261 &0.1801 &0.0897 &22.5482 \\
&&J&SSIM &0.3850 &0.4131 &0.2630 &2.2099  &0.2393 &0.2881 &0.1738 &21.9508 \\ \hdashline
\multirow{7}{*}{NR} &\multirow{3}{38pt}{{{Model-based}}} 
&K& 3D-NSS & 0.7144 & 0.7382  & 0.5174 & 1.7686    & 0.6479    & 0.6514    & 0.4417   & 16.5716 \\
&&L& ResSCNN &0.8600 & 0.8100 &- &- &- &- &- &-\\ 
&&M& GPA-Net &0.8750 & \bf\textcolor{blue}{0.8860} &- &- &0.7580 &0.7690 &- &-\\ \cdashline{2-12}

&\multirow{4}{38pt}{{{Projection-based}}} &N&
PQA-net & 0.8500 & 0.8200 & - & - & 0.7000 & 0.6900 & 0.5100 & 15.1800 \\
&&O&IT-PCQA & 0.5800 & 0.6300 & - & - & 0.5500 & 0.5400 & - & -\\
&&P&VQA-PC & 0.8509 & 0.8635 & 0.6585 & 1.1334 & \bf\textcolor{blue}{0.7968} & \bf\textcolor{blue}{0.7976} & \bf\textcolor{blue}{0.6115} & \bf\textcolor{blue}{13.6219}\\ \cdashline{3-12}
&&Q&\textbf{GMS-3DQA} &\bf\textcolor{red}{0.9108} &\bf\textcolor{red}{0.9177}	& \bf\textcolor{red}{0.7735}	& \bf\textcolor{red}{0.7872} & \bf\textcolor{red}{0.8308} & \bf\textcolor{red}{0.8338}	& \bf\textcolor{red}{0.6457}	& \bf\textcolor{red}{12.2292} 
 \\

\bottomrule
\end{tabular}}
\vspace{-0.3cm}
\label{tab:pcqa}
\end{table*}

\begin{table*}[!t]
\centering 
\renewcommand\tabcolsep{3.5pt}
\caption{Performance results on the CMDM and TMQ databases, where only 3D-NSS is model-based method. The best performance results are marked in {\bf\textcolor{red}{RED}} and the second performance results are marked in {\bf\textcolor{blue}{BLUE}}.}
\resizebox{\linewidth}{!}{\begin{tabular}{l:l:c:l|cccc|cccc}
\toprule
\multirow{2}{*}{Ref}&\multirow{2}{*}{Type}&\multirow{2}{*}{Index}&\multirow{2}{*}{Methods} & \multicolumn{4}{c|}{CMDM} & \multicolumn{4}{c}{TMQ}  \\ \cline{5-12}
        &&& & SRCC$\uparrow$      & PLCC$\uparrow$      & KRCC$\uparrow$     & RMSE $\downarrow$  & SRCC$\uparrow$      & PLCC$\uparrow$      & KRCC$\uparrow$     & RMSE $\downarrow$\\ \hline
\multirow{3}{*}{FR}&\multirow{3}{38pt}{{{Projection-based}}} &A&
PSNR &0.6129 &0.6557 &0.4902	&0.8665 &0.5295 &0.6535 &\bf\textcolor{blue}{0.3938} & \bf\textcolor{blue}{0.7877}\\
&&B&SSIM &0.7933 &0.8087 &0.6689		&0.5987 &0.4020 &0.5982 &0.2821 &0.8339 \\ 
&&C&G-LPIPS* &\bf\textcolor{red}{0.8900} & \bf\textcolor{red}{0.8800} &- &- &\bf\textcolor{red}{0.8600} & \bf\textcolor{red}{0.8500} &- &-\\ \hdashline
\multirow{4}{*}{NR}&Model-based&D&
3D-NSS &\bf\textcolor{blue}{0.8626} & 0.8754 & \bf\textcolor{red}{0.7222} & \bf\textcolor{blue}{0.6062} &0.4263	&0.4429	&0.2934	&1.0542  \\ \cdashline{2-12}
&\multirow{3}{38pt}{{{Projection-based}}}&E&BRISQUE &0.5295	&0.5906	&0.3688	&1.058 & 0.5364 & 0.4849  & 0.3788 & 0.9014    \\
&&F&NIQE &0.6694 &0.7505	&0.5342	&0.7491 & 0.3731 & 0.3866  & 0.2528 & 0.8782    \\ \cdashline{3-12}
&&G&\textbf{GMS-3DQA} & 0.8394 & \bf\textcolor{blue}{0.8759} & \bf\textcolor{blue}{0.6822} & \bf\textcolor{red}{0.5328} & \bf\textcolor{blue}{0.7810} & \bf\textcolor{blue}{0.7895} & \bf\textcolor{red}{0.5783} & \bf\textcolor{red}{0.5978} 
\\
\bottomrule
\end{tabular}}
\vspace{-0.3cm}
\label{tab:mqa}
\end{table*}

\subsection{Performance Discussion}
\label{sec:performance}
The experimental results are documented in Table \ref{tab:pcqa} and Table \ref{tab:mqa}, from which we can draw several conclusions: (a) As shown in Table \ref{tab:pcqa}, the proposed GMS-3DQA method outperforms all the comparison methods on the SJTU-PCQA and WPC databases, demonstrating the effectiveness of the proposed method on PCQA tasks. For example, GMS-3DQA is 3.7\% ahead of the second-place method GraphSIM (0.9108 vs. 0.8783) and is 4.2\% superior to the second-ranking NR method VQA-PC (0.8308 vs. 0.7968) on the SJTU-PCQA and WPC databases in terms of SRCC values respectively. (b) Additionally, a significant decline in performance was observed for all 3DQA methods when transitioning from the SJTU-PCQA database to the WPC database. This can be attributed to the more complex distortions and finer-grained degradation levels introduced by the WPC database. (c) In Table \ref{tab:mqa}, the proposed GMS-3DQA method is found to be inferior to the NR-MQA method 3D-NSS on the CMDM database and outperforms all the NR-MQA methods on the TMQ database. This might be because the CMDM database contains only 80 distorted meshes, which can result in overfitting due to insufficient training data. As a consequence, the performance of GMS-3DQA is inferior to that of 3D-NSS on the CMDM database. However, when applied to the TMQ database, which contains a larger and more complex set of distorted meshes, 3D-NSS experiences a significant performance decline, while the proposed method maintains a satisfactory level of performance. This observation suggests that the proposed model exhibits superior capability in adapting to and characterizing intricate distortions in 3D models. Moreover, it should be noted that G-LPIPS operates on projections from perceptually selected viewpoints, which limits its practical value compared to the proposed method. 

{ In conclusion, GMS-3DQA significantly boosts the performance of the NR-PCQA and NR-MQA methods. Given that GMS-3DQA relies on perpendicular projections, it's adaptable to other 3D digital formats. This adaptability facilitates cross-domain quality assessment across various 3D digital formats, potentially leading to a unified approach in 3D quality assessment. }

\begin{table}[!tp]\footnotesize
    \centering
    \caption{Illustration of flops, parameters, and average inference time (on CPU/GPU) per point cloud of the SJTU-PCQA and WPC databases. \textbf{Rendering time is included for the projection-based method and rendering one projection takes about 0.2s}. The subscript `$\mathbf{_{A\times}}$' of the consuming time indicates the corresponding method takes up $\mathbf{A\times}$ operation time of the proposed \textbf{GMS-3DQA}. }
    \vspace{0.1cm}
    \begin{tabular}{c|cc|c}
    \toprule
    {Method} & {Para. (M)} & {Gflops} & {Time (S) CPU/GPU} \\ \hline
        PCQM      & - & -  & 12.23{$\mathbf{_{7.83\times}}$}/-  \\
        GraphSIM  & - & - &270.14{$\mathbf{_{173.16\times}}$}/-  \\ 
       
        PointSSIM &- & -   &9.27{$\mathbf{_{5.94\times}}$}/-  \\
        3D-NSS    &- &-    &5.12\textcolor{blue}{$\mathbf{_{3.28\times}}$}/-   \\ 
        VQA-PC    &58.37 &50.08   &19.21{$\mathbf{_{12.31\times}}$}/16.44{$\mathbf{_{12.84\times}}$}   \\ \hdashline
        \textbf{GMS-3DQA} &27.54 & 4.38 &1.56\textcolor{red}{$\mathbf{_{1\times}}$}/1.28\textcolor{red}{$\mathbf{_{1\times}}$} \\ 
    \bottomrule
    \end{tabular}
    \label{tab:efficiency}
\end{table}

\begin{figure}[!t]
    \centering
    \includegraphics[width = 0.6\linewidth]{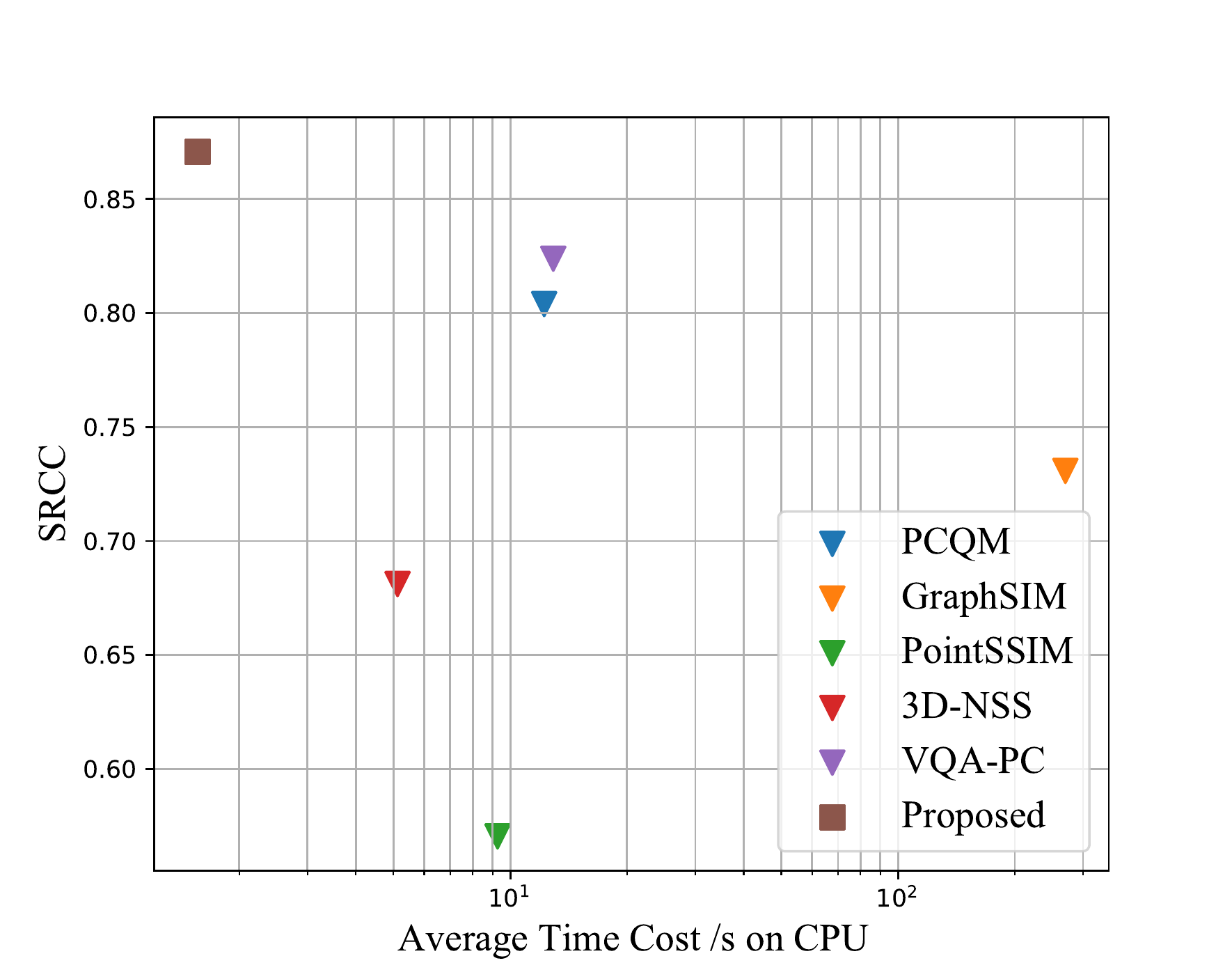}
    \caption{Comparison of the operating time on CPU, where the SRCC values are the average results of performance on the SJTU-PCQA and WPC databases. }
    \label{fig:efficiency}
\end{figure}

\subsection{Efficiency Analysis}
\label{sec:efficiency}
In the previous session, the effectiveness of GMS-3DQA is discussed while this section mainly focuses on the model efficiency.
The high efficiency of the proposed method mostly lies in turning the multi-projections into one QMM and inferring the perceptual quality from the single QMM input rather than multi-projections, which greatly reduces the inference time and computational complexity. A comparison for the computational complexity of 3DQA models is conducted with three FR-PCQA methods (PCQM, GraphSIM, and PointSSIM) and two NR-PCQA methods (3D-NSS and VQA-PC). It is noteworthy that both GMS-3DQA and VQA-PC are deep learning-based methods, while the remaining methods are handcrafted-based. The reason for excluding PQA-net is that PQA-net is trained on the non-public Waterloo Point Cloud Sub-Dataset (WPCSD) \cite{liu2021pqa}, which includes 7.920 distorted point clouds. The operation time is evaluated on a computer with Intel 12500H @ 3.11 GHz CPU, 16G RAM, and NVIDIA Geforce RTX 3070Ti GPU, running on the Windows operating system. The efficiency comparison results are presented in Table \ref{tab:efficiency} and Fig. \ref{fig:efficiency}. It can be observed that the proposed GMS-3DQA requires only 1/3.28 the inference time compared to the fastest competitor 3D-NSS while achieving the best performance on the CPU. Additionally, the proposed GMS-3DQA consumes half parameters and much fewer flops than the deep learning-based method VQA-PC. These comparisons further affirm the superior efficiency of the proposed GMS-3DQA method.

\begin{table*}[!tp]
\centering
\renewcommand\tabcolsep{2.4pt}
\caption{SRCC \& PLCC performance results of the ablation study. The efficiency information is listed as well.}
\resizebox{\linewidth}{!}{\begin{tabular}{c:c:c:c:c:c|cc|cc|cc|cc}
\toprule
  \multirow{2}{*}{Index} & \multirow{2}{*}{GMS} & \multirow{2}{*}{QMM} & \multirow{2}{*}{Para. (M)} & \multirow{2}{*}{Gflops} & Time (S)  & \multicolumn{2}{c|}{SJTU} & \multicolumn{2}{c|}{WPC} & \multicolumn{2}{c|}{CMDM} & \multicolumn{2}{c}{TMQ} \\ \cline{7-14}
  &&&&&CPU/GPU & SRCC$\uparrow$      & PLCC$\uparrow$  & SRCC$\uparrow$      & PLCC$\uparrow$  & SRCC$\uparrow$      & PLCC$\uparrow$ &SRCC$\uparrow$      & PLCC$\uparrow$  \\ \hline
 I&$\times$ &$\times$ &27.54 & 26.28 &2.51/1.67 & 0.8843 & 0.8861 & 0.8132 & 0.8222 & 0.8022 & 0.8127 & 0.7558 & 0.7563\\
II&$\checkmark$ & $\times$ &27.54 & 26.28 &2.44/1.66 &\bf\textcolor{blue}{0.9001} & \bf\textcolor{blue}{0.9002} & \bf\textcolor{blue}{0.8241} & \bf\textcolor{blue}{0.8244} & \bf\textcolor{blue}{0.8111} & \bf\textcolor{blue}{0.8441} & \bf\textcolor{blue}{0.7619} & \bf\textcolor{blue}{0.7633}\\
III&$\checkmark$ & $\checkmark$ &27.54 & 4.38 & 1.56/1.28 &\bf\textcolor{red}{0.9108} & \bf\textcolor{red}{0.9177} & \bf\textcolor{red}{0.8308} & \bf\textcolor{red}{0.8338} & \bf\textcolor{red}{0.8394} & \bf\textcolor{red}{0.8759} & \bf\textcolor{red}{0.7810} & \bf\textcolor{red}{0.7895}\\ 

\bottomrule
\end{tabular}}
\label{tab:ablation}
\end{table*}

\subsection{Ablation Study}
{To fully investigate the contribution of the multi-projection grid mini-patch sampling (GMS-3DQA) strategy, we conduct the ablation studies from two aspects: (a) We evaluate the performance of the bare model without GMS and QMM. More specifically, the six perpendicular projections are resized and cropped into 224$\times$224 resolutions and then put into the same ST-t backbone feature extraction. Then extracted features are fused with average pooling and the default quality regression module is employed to predict the quality levels. (b) We evaluate the model's performance (using GMS but excluding QMM) in this part. In this scenario, we use 6 grid mini-patch maps sampled from the 6 projections as input. This is followed by the processes of feature extraction, feature fusion, and quality regression, all conducted as previously detailed.}

The experimental results are exhibited in Table \ref{tab:ablation}. With closer inspections, we can draw several useful conclusions: (a) The model equipped with GMS and QMM achieves the best performance, which indicates both GMS and QMM makes contributions to the final results. (b) The GMS strategy helps the model gain better performance than the bare model with resize and crop strategy, which indicates the GMS can better preserve the quality-aware patterns from the rendered projections. (c) The QMM further improves performance. It is because the existing quality information redundancy within the 6 grid mini-patch maps may confuse the model and leads to a performance drop. The QMM actively integrates the quality information from the viewpoints and reduces the information redundancy within each grid mini-patch map, which assists the model to gain a better perceptual understanding of the 3D model. (d) Furthermore, the utilization of QMM only needs 
extracting features from a single quality map while models without QMM need to extract features from all the projections, which requires more computational resources as well as inference time.

\begin{figure*}[!t]
    \centering
    \subfigure[SJTU-PCQA]{\begin{minipage}[t]{.45\linewidth}
                \centering
                \includegraphics[width = \linewidth]{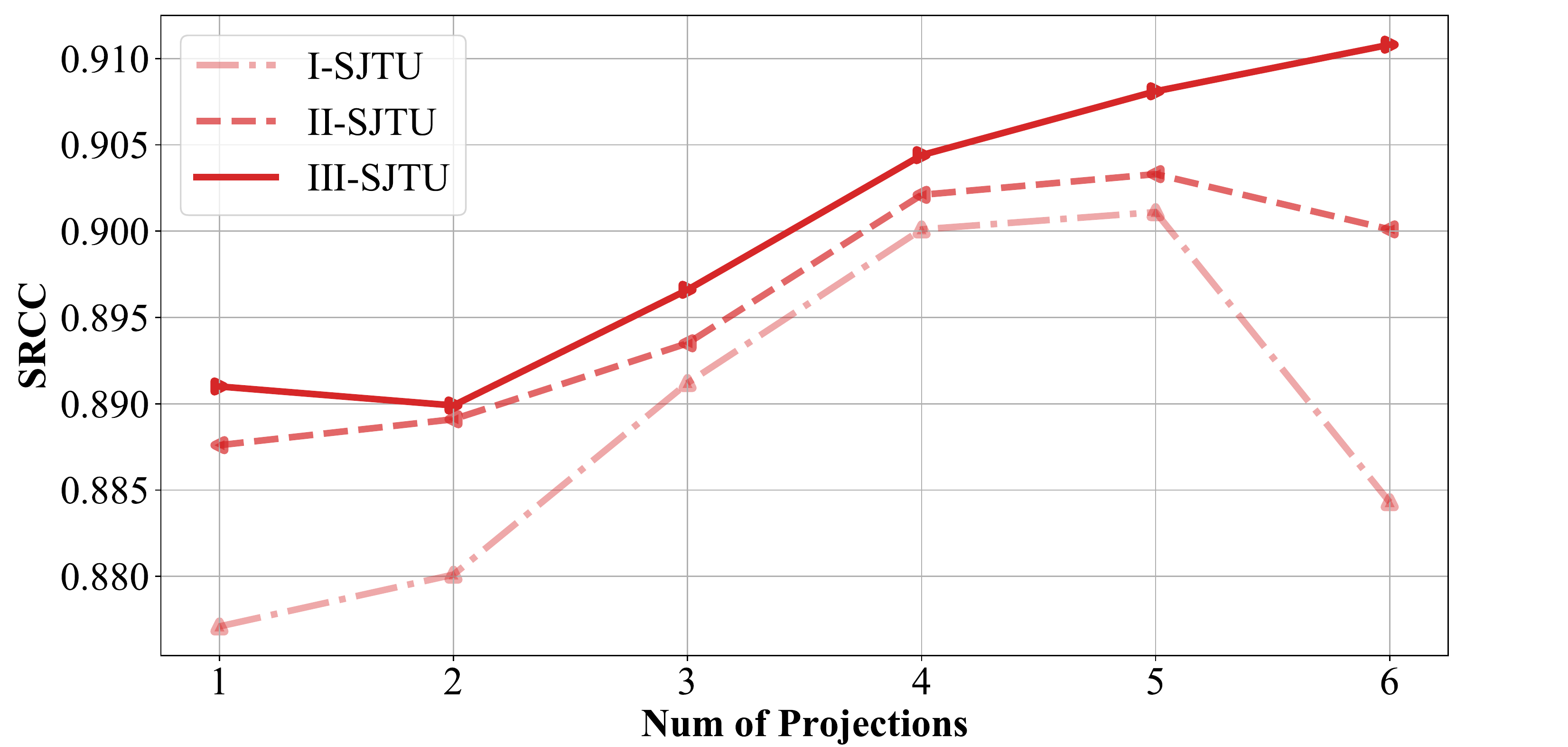}
                \end{minipage}}
    \subfigure[WPC]{\begin{minipage}[t]{.45\linewidth}
                \centering
                \includegraphics[width = \linewidth]{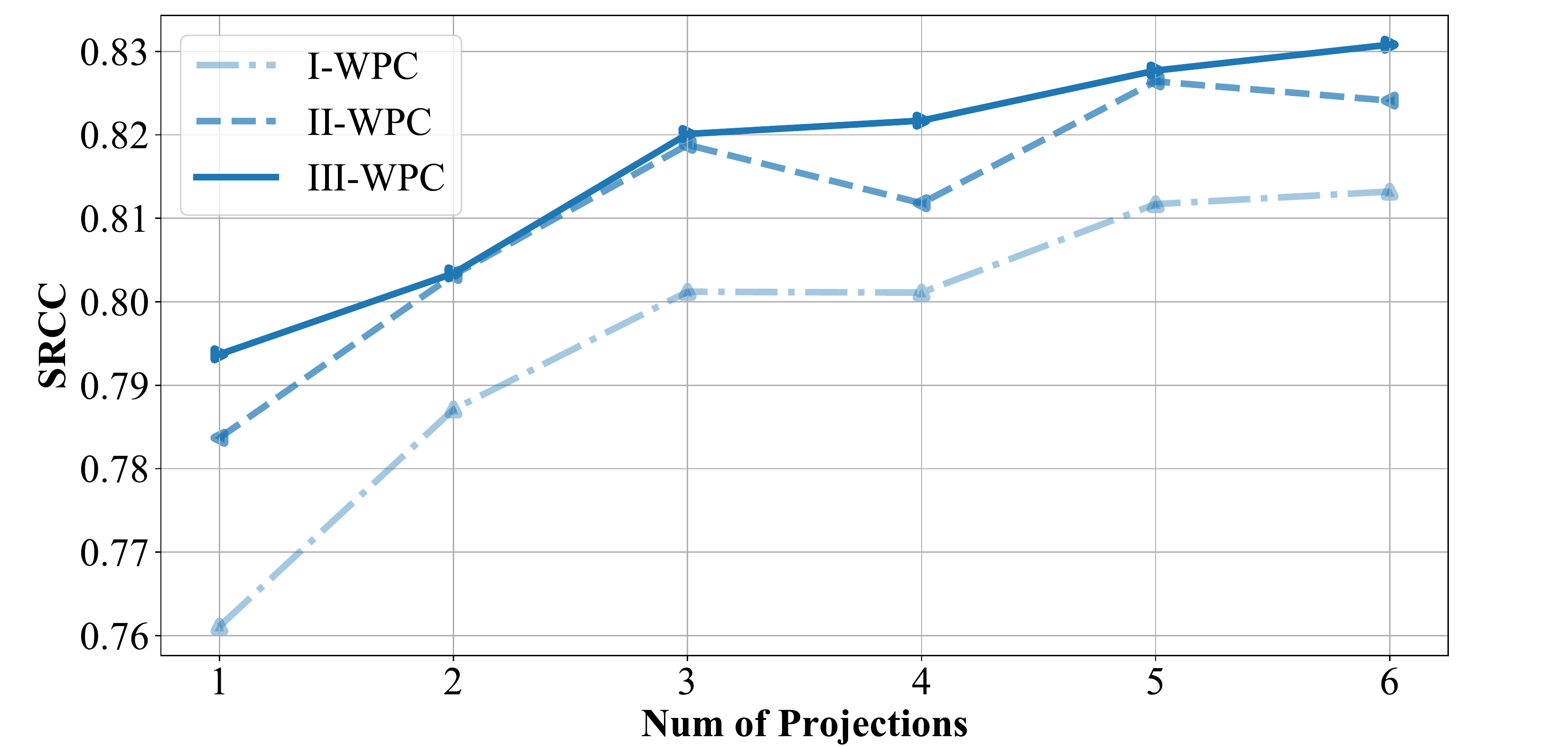}
                \end{minipage}}
    \subfigure[CMDM]{\begin{minipage}[t]{.45\linewidth}
                \centering
                \includegraphics[width = \linewidth]{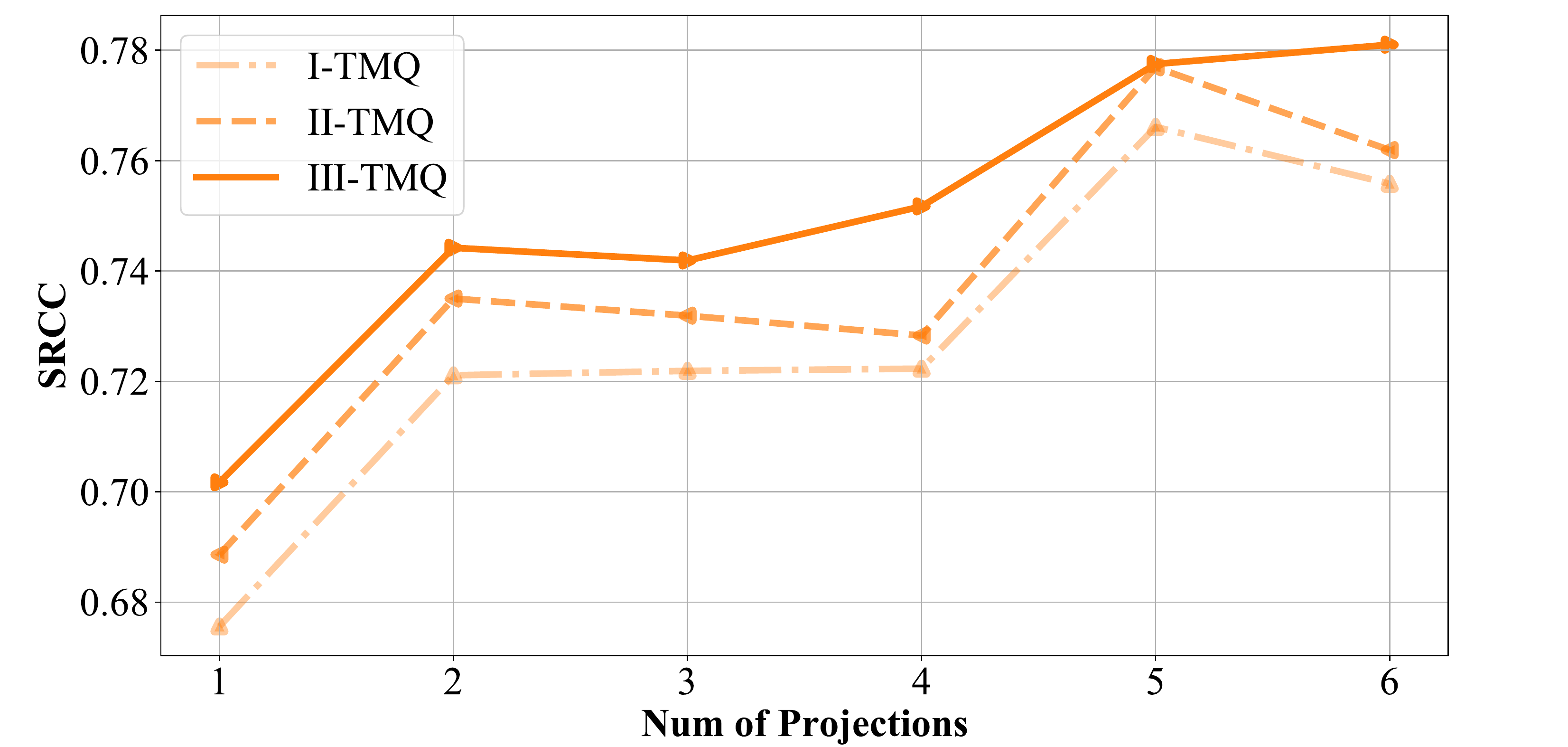}
                \end{minipage}}
    \subfigure[TMQ]{\begin{minipage}[t]{.45\linewidth}
                \centering
                \includegraphics[width = \linewidth]{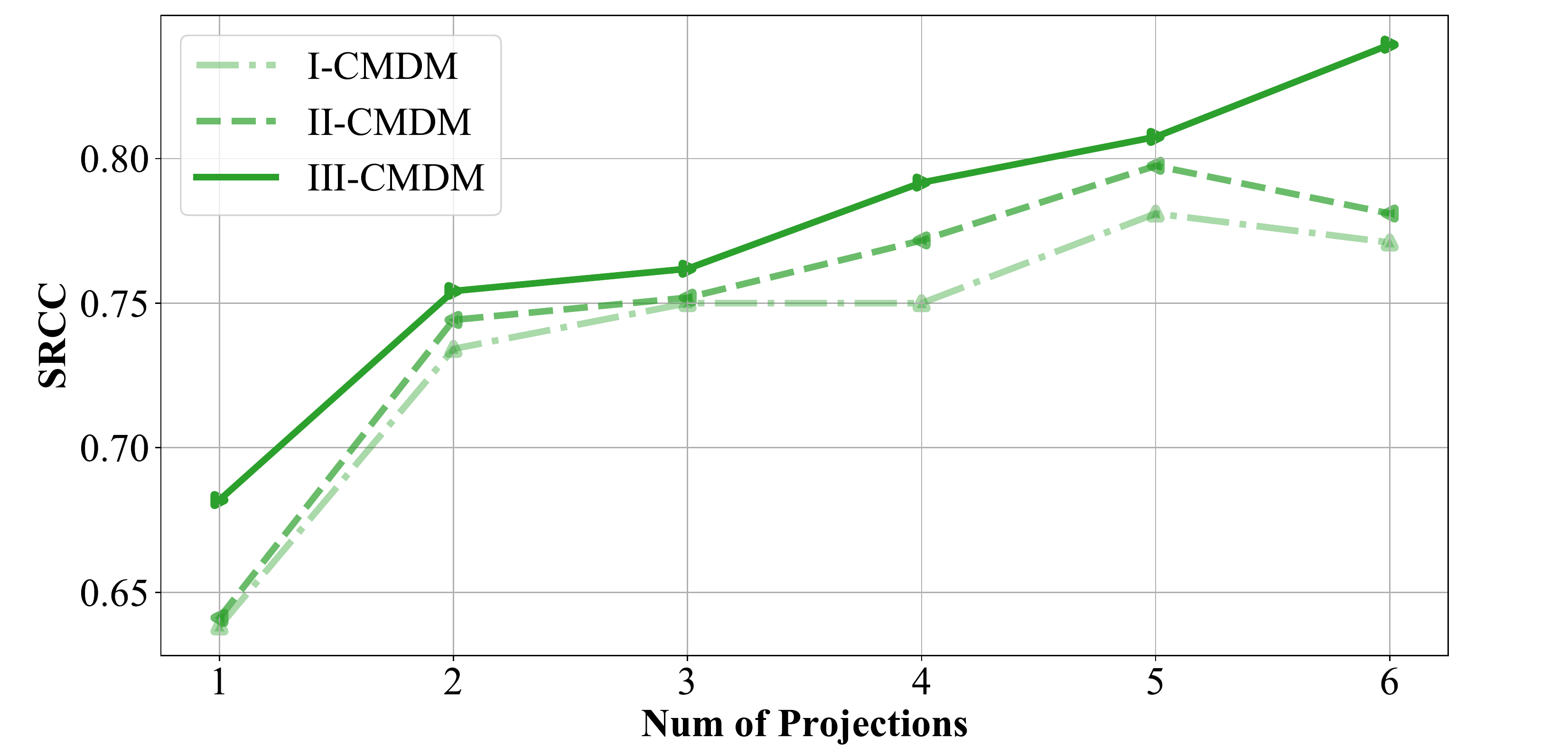}
                \end{minipage}}

    \caption{Illustration of the SRCC performance corresponding to the number of projections, where the label is represented in the format of `Index-Database' and the index is organized as the same order in Table \ref{tab:ablation}. For example, `I-SJTU' represent the SRCC performance of model I on the SJTU-PCQA database. }
    \label{fig:number}
\end{figure*}

\subsection{Effect of the Number of Projections}
In this section, we investigate the impact of the number of projections on the performance of the proposed method. Our results, depicted in Figure \ref{fig:number}, show that all six viewpoints contribute significantly to the performance improvement of the proposed method. Interestingly, our analysis also reveals that increasing the number of projections from 5 to 6 leads to a noticeable decrease in the performance of Model I and Model II on all but one of the four tested databases, namely `I-WPC'. However, Model III demonstrates a positive trend in performance with increasing numbers of projections, and when the number of projections increases from 5 to 6, the performance of the proposed method is further improved. These findings suggest that the proposed method effectively leverages and integrates quality-aware information from multi-projections while minimizing the negative impact of redundancy.

\begin{table*}[!tp]
\centering
\renewcommand\tabcolsep{2.4pt}
\caption{SRCC \& PLCC performance results for utilizing different backbones. The efficiency information is listed as well.}
\resizebox{\linewidth}{!}{\begin{tabular}{c:c:c:c|cc|cc|cc|cc}
\toprule
  \multirow{2}{*}{Backbone} & \multirow{2}{*}{Para. (M)} & \multirow{2}{*}{Gflops} & Time (S)  & \multicolumn{2}{c|}{SJTU} & \multicolumn{2}{c|}{WPC} & \multicolumn{2}{c|}{CMDM} & \multicolumn{2}{c}{TMQ} \\ \cline{5-12}
  &&&CPU/GPU & SRCC$\uparrow$      & PLCC$\uparrow$  & SRCC$\uparrow$      & PLCC$\uparrow$  & SRCC$\uparrow$      & PLCC$\uparrow$ &SRCC$\uparrow$      & PLCC$\uparrow$  \\ \hline
 ResNet50 &25.60 & 4.11 &1.60/1.30 & 0.9030 & 0.9016 & \bf\textcolor{blue}{0.8114} & \bf\textcolor{blue}{0.8111} & 0.8174 & 0.8172 & 0.7705 & 0.7637\\
 ConvNeXtV2-Tiny&27.84 & 4.45 &1.59/1.28 &\bf\textcolor{blue}{0.9101} & \bf\textcolor{blue}{0.9100} & {0.7682} & {0.7610} & \bf\textcolor{blue}{0.8294} & \bf\textcolor{blue}{0.8319} & \bf\textcolor{blue}{0.7766} & \bf\textcolor{blue}{0.7788}\\
 Swin-Trans. tiny&27.54 & 4.38 & 1.56/1.28 &\bf\textcolor{red}{0.9108} & \bf\textcolor{red}{0.9177} & \bf\textcolor{red}{0.8308} & \bf\textcolor{red}{0.8338} & \bf\textcolor{red}{0.8394} & \bf\textcolor{red}{0.8759} & \bf\textcolor{red}{0.7810} & \bf\textcolor{red}{0.7895}\\ 

\bottomrule
\end{tabular}}
\label{tab:backbone}
\end{table*}

\subsection{Backbone Comparison}
To explore the impact of different backbones on GMS-3DQA, we conduct a comparative experiment using two backbone networks, ResNet50 \cite{he2016deep} and ConvNeXtV2-Tiny \cite{woo2023convnext}, which have similar parameter counts and FLOPs as ST-t. Both ResNet50 and ConvNeXtV2-Tiny are initialized with the weights pretrained on the ImageNet-22K database \cite{deng2009imagenet}. The experimental results are exhibited in Table \ref{tab:ablation}. Substituting different backbones does not severely affect the performance of GMS-3DQA, thus demonstrating its robustness. Further investigations reaveal that the ST-t backbone achieves the best performance, indicating that ST-t is the optimal choice for this task.

\begin{table*}[t]
    \renewcommand\tabcolsep{12pt}
    \centering
    \caption{Performnace results for cross-database evaluation, where WPC$\rightarrow$SJTU-PCQA indicates the model is trained on the WPC database and validated with the default testing setup of the SJTU database. }
    \resizebox{\linewidth}{!}{\begin{tabular}{c|cc|cc|cc|cc}
    \toprule
    \multirow{2}{*}{Model}  & \multicolumn{2}{c|}{WPC$\rightarrow$SJTU} & \multicolumn{2}{c|}{SJTU$\rightarrow$WPC} & \multicolumn{2}{c|}{TMQ$\rightarrow$CMDM} &  \multicolumn{2}{c}{CMDM$\rightarrow$TMQ} \\ \cline{2-9}
            & SRCC$\uparrow$    & PLCC$\uparrow$   & SRCC$\uparrow$   & PLCC$\uparrow$ & SRCC$\uparrow$    & PLCC$\uparrow$   & SRCC$\uparrow$   & PLCC$\uparrow$\\ \hline
        PQA-net       & {0.5411} & {0.6102} & - & - & - & - & - & -\\ 
        3D-NSS        & 0.1817 & 0.2344 & 0.1512 & 0.1422 & - & - & - & -\\ 
        VQA-PC        & \bf\textcolor{blue}{0.7172} & \bf\textcolor{blue}{0.7233} & \bf\textcolor{blue}{0.2212} & \bf\textcolor{blue}{0.2317} & - & - & - & -\\\hdashline
        BRISQUE        & 0.5148 & 0.4907 & 0.2158 & 0.2198 & \bf\textcolor{blue}{0.4138} & \bf\textcolor{blue}{0.4511} & \bf\textcolor{blue}{0.0140} & \bf\textcolor{blue}{0.0157}\\ \hdashline
       \bf{GMS-3DQA}       & \bf\textcolor{red}{0.7421}   & \bf\textcolor{red}{0.7611}    &\bf\textcolor{red}{0.3196}    & \bf\textcolor{red}{0.3321} & \bf\textcolor{red}{0.7921}   & \bf\textcolor{red}{0.8123}    &\bf\textcolor{red}{0.2597}    & \bf\textcolor{red}{0.2992}\\     
    \bottomrule
    \end{tabular}}
    \label{tab:crossdatabase}
    \vspace{-0.2cm}
\end{table*}

\subsection{Cross-Database \& Cross-Domain Validation}
\label{sec:cross}
In this section, we aim to assess the generalization capability of the proposed method by conducting a cross-database experiment. Specifically, we investigate the potential for cross-domain generalization, which refers to the ability of the learned quality representation from point clouds to be adapted to mesh and vice versa. It's worth mentioning that the default $k$-fold testing sets as well as the implementation setup are used so that the generalization performance can be directly compared with other methods' performance in Table \ref{tab:pcqa} and Table \ref{tab:mqa}. The cross-database performance is reported in Table \ref{tab:crossdatabase}. The proposed GMS-3DQA presents the best generalization performance among the NR-3DQA competitors. The unsatisfied SJTU$\rightarrow$WPC (Scale: 378$\rightarrow$740) and CMDM$\rightarrow$TMQ (Scale: 80$\rightarrow$3000) performance is due to the over-fitting resulting from the limited scale and distortion types of the SJTU-PCQA ND CMDM databases.

Then we try to conduct the cross-domain validation. More specifically, we test the PCQA performance with the model trained on the MQA database and vice versa. \textbf{Since nearly all the competitors are targeted at only one domain (point cloud or mesh), we list the intra-database performance of some NR competitors for exhibition.} The cross-domain results can be found in Table \ref{tab:crossdomain}, from which we can make several observations: (a) GMS-3DQA's performance is significantly enhanced when trained on databases with larger scales and more complex distortions. For instance, GMS-3DQA trained on the WPC database achieves better performance on the CMDM (SRCC: 0.49 vs. 0.60) and WPC (SRCC: 0.28 vs. 0.53) databases than SJTU, while GMS-3DQA trained on the TMQ database achieves better results on the SJTU (SRCC: 0.62 vs. 0.75) and WPC (SRCC: 0.40 vs. 0.60) databases than CMDM. (b) It's interesting to find that GMS-3DQA trained on TMQ achieves an acceptable gap with the best intra-database-trained competitor VQA-PC on the SJTU (SRCC: 0.75 vs. 0.85) and WPC (SRCC: 0.60 vs. 0.70) databases, which indicates GMS-3DQA is capable of modeling the shared perceptual representations regardless of the digital formats and gains competitive domain generalization ability.  

In conclusion, the proposed method gains the potential to deal with the QA tasks on a target 3D domain containing insufficient annotated data by using the knowledge learned from the other related 3D domain with adequate labeled data.

\begin{table*}[t]
\renewcommand\tabcolsep{2pt}
    \centering
    \caption{Performance results for cross-domain evaluation. The scale for each database is marked. Since nearly all the competitors are targeted at only one domain (point cloud or mesh), we list the intra-database performance of some NR competitors for exhibition, which indicates that {\bf{only the proposed GMS-3DQA is validated in the cross-domain format and other NR competitors are validated in the intra-database* format.}} }
    \resizebox{\linewidth}{!}{\begin{tabular}{c:c|cc:cc|cc:cc|cc:cc|cc:cc}
    \toprule
    \multirow{3}{*}{Type} & \multirow{3}{*}{Model} & \multicolumn{8}{c|}{PCQA$\rightarrow$MQA (SJTU:378, WPC:740)} & \multicolumn{8}{c}{MQA$\rightarrow$PCQA (CMDM:80, WPC:3,000)} \\ \cline{3-18}
   & & \multicolumn{2}{c|}{SJTU$\rightarrow$CMDM} & \multicolumn{2}{c|}{WPC$\rightarrow$CMDM} & \multicolumn{2}{c|}{SJTU$\rightarrow$TMQ} & \multicolumn{2}{c|}{WPC$\rightarrow$TMQ} & \multicolumn{2}{c|}{CMDM$\rightarrow$SJTU} & \multicolumn{2}{c|}{TMQ$\rightarrow$SJTU} & \multicolumn{2}{c|}{CMDM$\rightarrow$WPC}  & \multicolumn{2}{c}{TMQ$\rightarrow$WPC}\\ \cline{3-18}
   & & SRCC    & PLCC   & SRCC   & PLCC & SRCC    & PLCC   & SRCC   & PLCC & SRCC    & PLCC   & SRCC   & PLCC & SRCC    & PLCC   & SRCC   & PLCC \\ \hline
      \multirow{3}{*}{Intra*} & PQA-net*  & - & - & - & - & - & - & - & - & \bf\textcolor{blue}{0.85} & \bf\textcolor{blue}{0.82} & \bf\textcolor{blue}{0.85} & \bf\textcolor{blue}{0.82} &\bf\textcolor{blue}{0.70} & \bf\textcolor{blue}{0.69} &\bf\textcolor{blue}{0.70} & \bf\textcolor{blue}{0.69} \\
       &  VQA-PC*  & - & - & - & - & - & - & - & - & \bf\textcolor{red}{0.85} & \bf\textcolor{red}{0.86} & \bf\textcolor{red}{0.85} & \bf\textcolor{red}{0.86} &\bf\textcolor{red}{0.79} &\bf\textcolor{red}{0.79} &\bf\textcolor{red}{0.79} &\bf\textcolor{red}{0.79} \\
       &  3D-NSS* &\bf\textcolor{red}{0.86} & \bf\textcolor{red}{0.87} & \bf\textcolor{red}{0.86} & \bf\textcolor{red}{0.87} & \bf\textcolor{red}{0.42} & \bf\textcolor{red}{0.42} & \bf\textcolor{blue}{0.42} & \bf\textcolor{blue}{0.42} & 0.71 & 0.73 & 0.71 &0.73 & 0.64 & 0.65 & 0.64 & 0.65\\  \hdashline
       Cross & \bf{GMS-3DQA} &\bf\textcolor{blue}{0.49} &\bf\textcolor{blue}{0.52} & \bf\textcolor{blue}{0.60} & \bf\textcolor{blue}{0.63} & \bf\textcolor{blue}{0.28} & \bf\textcolor{blue}{0.34} & \bf\textcolor{red}{0.53} & \bf\textcolor{red}{0.57} & 0.62 & 0.62 & 0.75 & 0.77 &0.40 &0.46  & 0.60 & 0.62\\     
    \bottomrule
    \end{tabular}}
    \label{tab:crossdomain}
    \vspace{-0.2cm}
\end{table*}

\begin{figure*}[!t]
    \centering
    \subfigure[SJTU-PCQA]{\begin{minipage}[t]{0.24\linewidth}
                \centering
                \includegraphics[width = \linewidth]{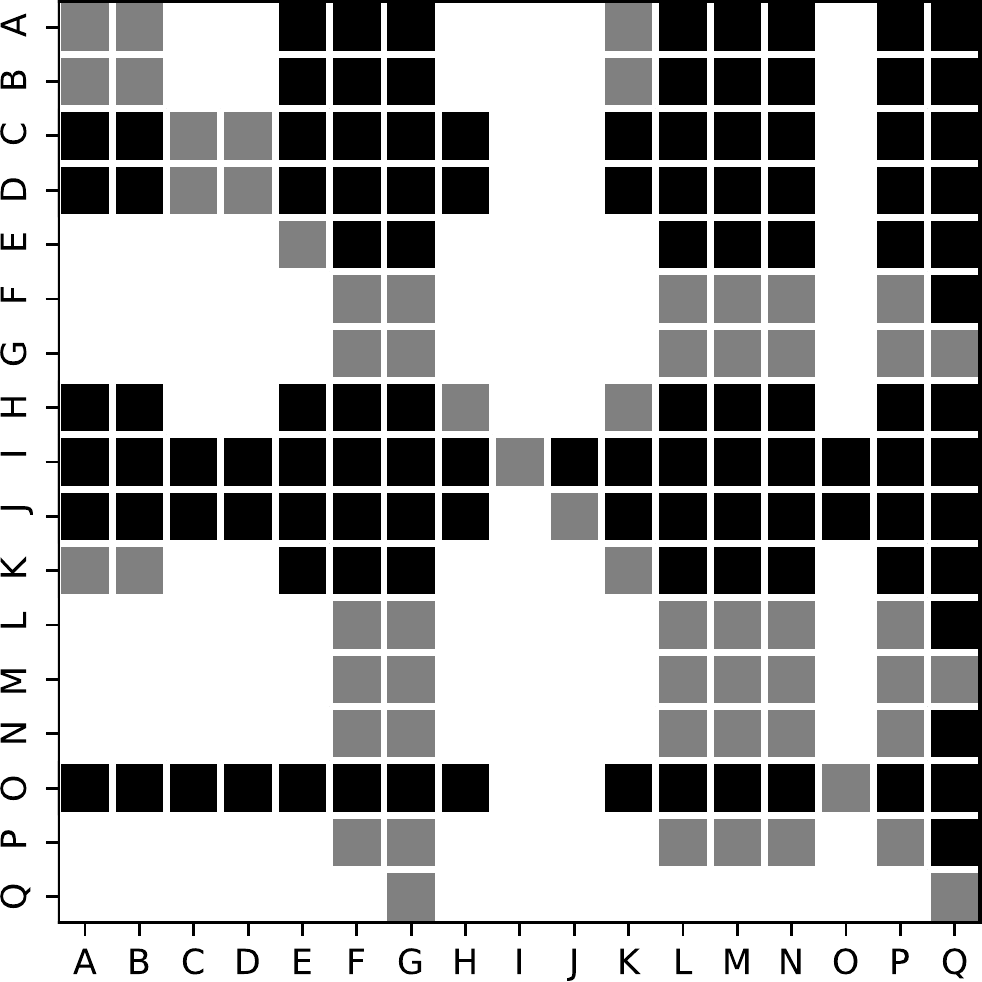}
                \end{minipage}}
    \subfigure[WPC]{\begin{minipage}[t]{0.24\linewidth}
                \centering
                \includegraphics[width = \linewidth]{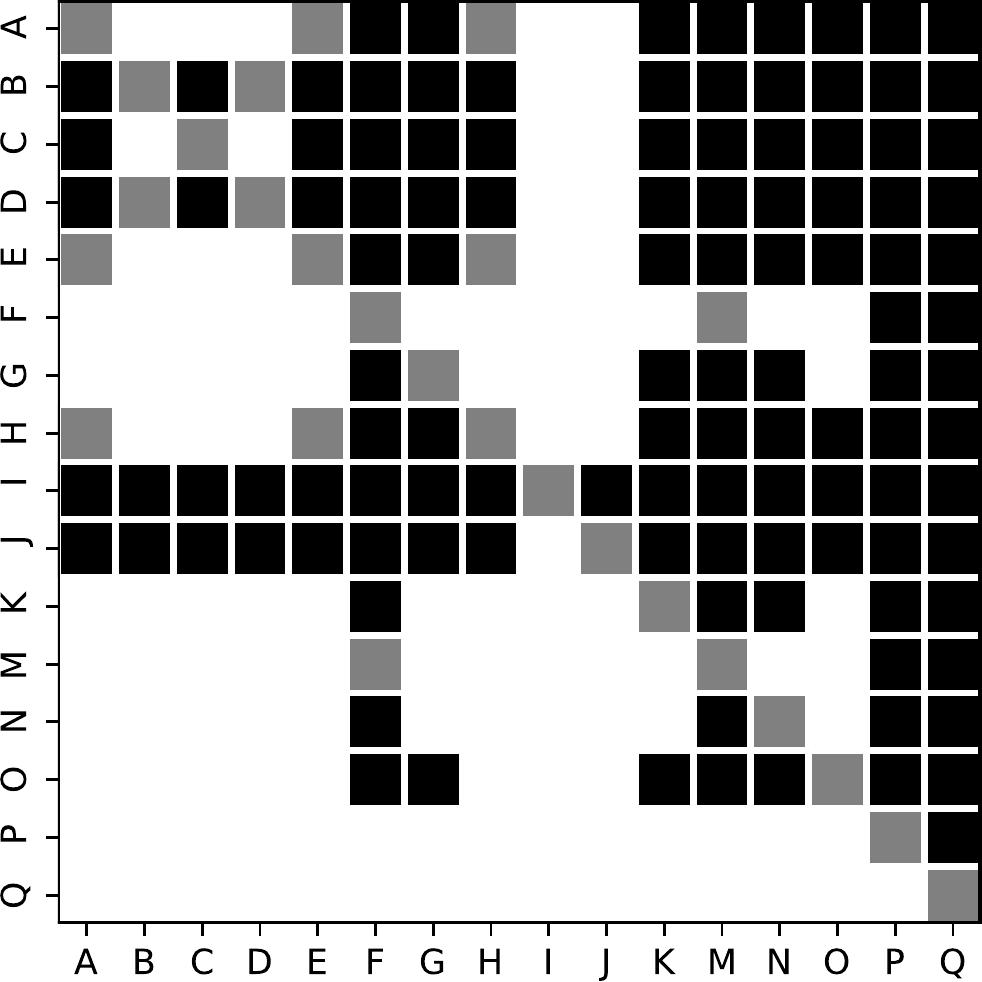}
                \end{minipage}}
    \subfigure[CMDM]{\begin{minipage}[t]{0.24\linewidth}
                \centering
                \includegraphics[width = \linewidth]{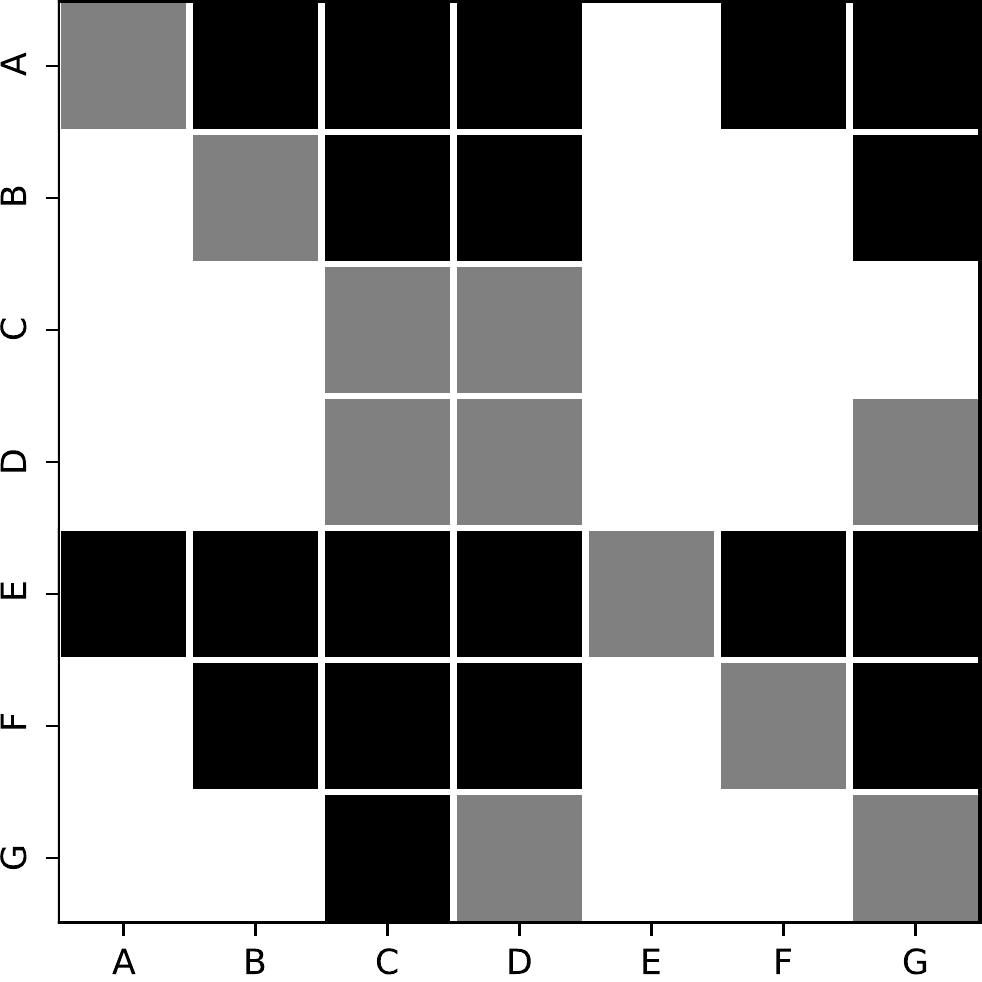}
                \end{minipage}}
    \subfigure[TMQ]{\begin{minipage}[t]{0.24\linewidth}
                \centering
                \includegraphics[width = \linewidth]{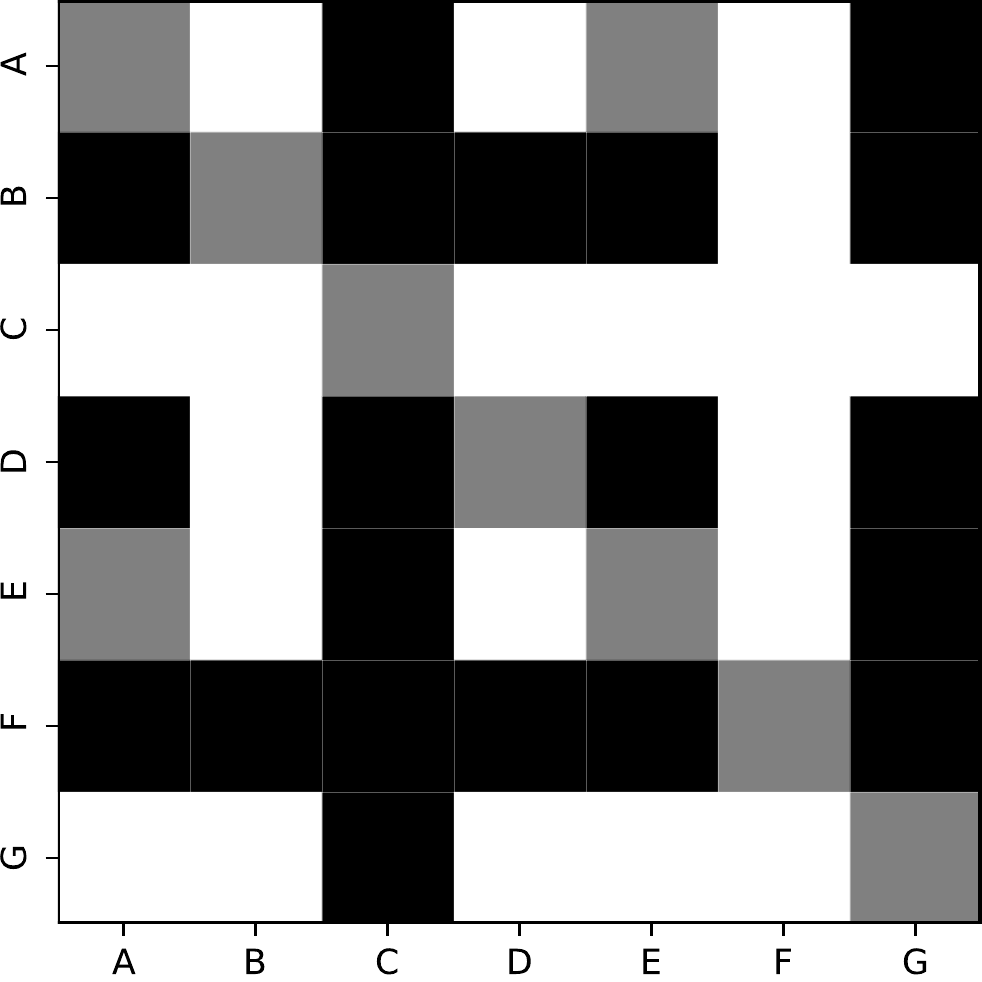}
                \end{minipage}}
    \caption{{The results of statistical tests on the SJTU-PCQA, WPC, CMDM, and TMQ databases. A black/white block indicates that the row method is inferior/superior to the column method, while a gray block signifies that there is no statistical difference between the row and column methods. The methods are identified by the same index as in Table \ref{tab:pcqa} and Table \ref{tab:mqa}.} }
    \label{fig:heatmap}
\end{figure*}

\subsection{Statistical Test}
{
To gain further insight into the performance of the proposed method, we conduct a statistical test in this section. Our experiment setup follows the same procedure outlined in \cite{statistic-test} and evaluates the difference between the predicted quality scores and the subjective ratings. All possible pairs of models are tested and the results are displayed in Fig. \ref{fig:heatmap}.
The results reveal that our method is significantly better than 14 and 15 PCQA metrics on the SJTU-PCQA and WPC databases respectively. Moreover, the proposed method significantly outperforms 4 and 5 compared MQA metrics on the CMDM and TMQ databases respectively. Notably, the proposed NR method is even not statistically distinguishable from the FR methods on the PCQA databases, which further confirms the proposed method's effectiveness.}

\section{Conclusion}
\label{sec:conclusion}
This paper presents a projection-based Grid Mini-patch Sampling 3D Model Quality Assessment (GMS-3DQA) method that aims to reconcile the core contradiction of utilizing multi-projection information while reducing the operation time. The method adopts a common rendering setup of six perpendicular viewpoints for rendering projections, which reduces the time spent on viewpoint selection and covers sufficient quality information. The projections are split into grid mini-patch maps and a single quality mini-patch map (QMM) is generated by randomly sampling mini-patches from each viewpoint. The use of QMM improves efficiency and aggregates quality information across different viewpoints, thus enhancing performance and robustness. The proposed GMS-3DQA method is validated on both PCQA and MQA databases, and outperforms all the compared NR-3DQA methods except on the CMDM database while being competitive with FR-3DQA methods. The results show that GMS-3DQA is 3.28 times faster than the fastest comparing 3DQA method on CPU and has fewer parameters and FLOPs.

\bibliographystyle{ACM-Reference-Format}
\bibliography{sample-base}










\end{document}